\documentclass{article}
\usepackage[a4paper]{geometry}
\usepackage{amsfonts,amsmath,amssymb} 
\usepackage{mathtools}
\usepackage{url}
\usepackage{geometry}

\usepackage{rotating}

\usepackage{bm}
\usepackage{eurosym}
\usepackage{graphicx}
\usepackage[colorlinks=true,pdftex=true]{hyperref} 
\usepackage[pagewise]{lineno}
\usepackage{listings}
\usepackage{multirow}
\usepackage{url}
\usepackage{xcolor}
\usepackage{xspace}
\usepackage[english]{babel}
\usepackage{lscape}
\usepackage{textcomp}
\usepackage{multirow}
\usepackage{stmaryrd}
\usepackage[ruled]{algorithm}
\usepackage[noend]{algpseudocode}
\usepackage{pbox}
\usepackage{listings}
\usepackage{verbatim}
\usepackage{tikz}
\usetikzlibrary{trees}
\usetikzlibrary{shadows}
\usetikzlibrary{arrows}
\usepackage{rotating,makecell}

\newcommand*{\qed}{\null\nobreak\hfill\ensuremath{\square}}

\floatname{algorithm}{Algorithm}

\newcommand{\ceil}[1]{\lceil #1 \rceil}

\newcommand{\nd}{\noindent}
\newcommand{\ind}{\mathsf{I}}
\newcommand{\indkb}{\ind_{\KB}}

\newcommand{\inds}[1]{\ind_{#1}}

\newcommand{\reffig}[1]{Figure~\ref{#1}}

\newcommand{\refalgo}[1]{Algorithm~\ref{#1}}

\newtheorem{remark}{Remark}

\newtheorem{example}{Example}[section]

\graphicspath{{./}{images/}}
\newcommand{\m}{d}

\newcommand{\unit}{[0,1]}
\newcommand{\unitp}{(0,1]}

\newcommand{\dt}{{\mathbf{d}}}

\newcommand{\bed}[2]{bed(#1, #2)}

\newcommand{\DL}{\ensuremath{\mathcal{DL}}}

\newcommand{\K}{\ensuremath{\mathcal{K}}}
\newcommand{\I}{\ensuremath{\mathcal{I}}\xspace}      

\newcommand{\EL}{\ensuremath{\mathcal{EL}}}

\newcommand{\ELW}{\ensuremath{\mathcal{ELW}}}
\newcommand{\ELNW}{\ensuremath{\mathcal{ELW^\neg}}}

\newcommand{\dllite}{\mbox{DL-Lite}}

\newcommand{\D}{{\mathbf{D}}}
\newcommand{\eqs}{\,{=}\,}

\newcommand{\calS}{{\cal S}}
\newcommand{\E}{{\cal E}}

\newcommand{\alc}{\mathcal{ALC}}

\newcommand{\el}{\mathcal{EL}}
\newcommand{\elbool}{\el(\D^-)}

\newcommand{\andc}{\sqcap}

\newcommand{\some}{\exists}

\newcommand{\notc}{\neg}

\newcommand{\csome}{\exists}

\newcommand{\bottomc}{\perp}
\newcommand{\topc}{\top}
\newcommand{\impc}{\sqsubseteq}
\newcommand{\highi}[1]{{#1}^{\cal I} }

\newcommand{\calC}{\ensuremath{\mathcal{C}} }
\newcommand{\calE}{\ensuremath{\mathcal{E}} }

\newcommand{\tuple}[1]{\langle #1 \rangle }

\newcommand{\fuzzyg}[2]{\mbox{$\tuple{#1,#2}$}}

\newcommand{\foil}{\textsc{Foil}}
\newcommand{\pfoil}{p\textsc{Foil}}

\newcommand{\foildl}{\foil-\DL}
\newcommand{\foildlw}{w\foildl}
\newcommand{\pfoildl}{\pfoil-\DL}
\newcommand{\fuzzyowladaboost}{\textsc{Fuzzy OWL-Boost}}

\algblockdefx[function]{FUNCTION}{ENDFUNCTION}[2]{\textbf{function} \texttt{#1(#2)}}{\textbf{end}}
\algblockdefx[returnCondition]{IfReturn}{EndIfReturn}[1]{\textbf{if}(#1)}
[1]{\indent \textbf{return} #1;}

\newcommand{\myvec}[1]{\mathbf{#1}}
\newcommand{\ie}{{\em i.e.}}
\newcommand{\eg}{{\em e.g.}}
\newcommand{\cf}{{\em cf.}}
\newcommand{\viz}{{\em viz.}}
\newcommand{\wrt}{{w.r.t.}}

\newcommand{\cass}[2]{\mbox{$#1$:$#2$}}
\newcommand{\rass}[3]{\mbox{$(#1,#2)$:$#3$}}

\newcommand{\ii}[1]{\emph{(#1)}}

\newcommand{\KB}{{\K}}
\newcommand{\fKB}{{\tilde{\K}}}

\title{\fuzzyowladaboost: Learning Fuzzy Concept Inclusions via Real-Valued Boosting}

\author{Franco Alberto Cardillo \\
ILC - CNR \\
Pisa, Italy \\
\and Umberto Straccia\\
ISTI - CNR \\
Pisa, Italy 
}

\begin{document}
\maketitle

%
%
%


\begin{abstract}
OWL ontologies are nowadays a quite popular way to describe structured knowledge in terms of classes, relations among classes and class instances.  

In this paper, given an OWL ontology and  a target class $T$,  we address the problem of learning fuzzy concept inclusion  axioms that describe sufficient conditions for being an individual instance of  $T$ (and to which degree). To do so, we present \fuzzyowladaboost~that relies on the $\mathbb{R}$eal AdaBoost boosting algorithm adapted to the (fuzzy) OWL case. We illustrate its effectiveness by means of an experimentation with several ontologies.
\end{abstract}


\section{Introduction}
\nd OWL 2 ontologies~\cite{OWL2} are nowadays a popular means to represent \emph{structured} knowledge and its formal semantics is based on \emph{Description Logics} (DLs)~\cite{Baader07a}. The basic ingredients of DLs are concept descriptions (in First-Order Logic terminology, unary predicates), inheritance relationships among them and instances of them.

Although an important amount of work has been carried about DLs, the application of machine learning techniques to OWL 2 ontologies, \viz~DL ontologies,  is relatively less addressed compared to the  \emph{Inductive Logic Programming} (ILP) setting (see \eg~\cite{deRaedt08,DeRaedt17} for more insights on ILP). We refer the reader to~\cite{Lisi19,Rettinger12} for an  overview.

In this work, we  focus on the problem of automatically learning fuzzy concept inclusion axioms from (crisp) OWL 2 ontologies. More specifically, given a  target class $T$ of an OWL ontology, we address the problem of learning fuzzy concept inclusion axioms that describe sufficient conditions for being an individual instance  of $T$ (and to which degree). An example illustrating the problem is shown next.

\begin{example}[Running example~\cite{Lisi13a,Lisi15,Straccia15}] \label{runes}
Consider an ontology that describes the meaningful entities of a city, such as \eg~the Hotel ontology in our experiments  (see Section~\ref{sec:eval}, Table~\ref{tab:onto}).
An excerpt of this ontology is given in Fig.~\ref{hotelonto}.
\begin{figure}
\begin{center}
\includegraphics[scale=0.325]{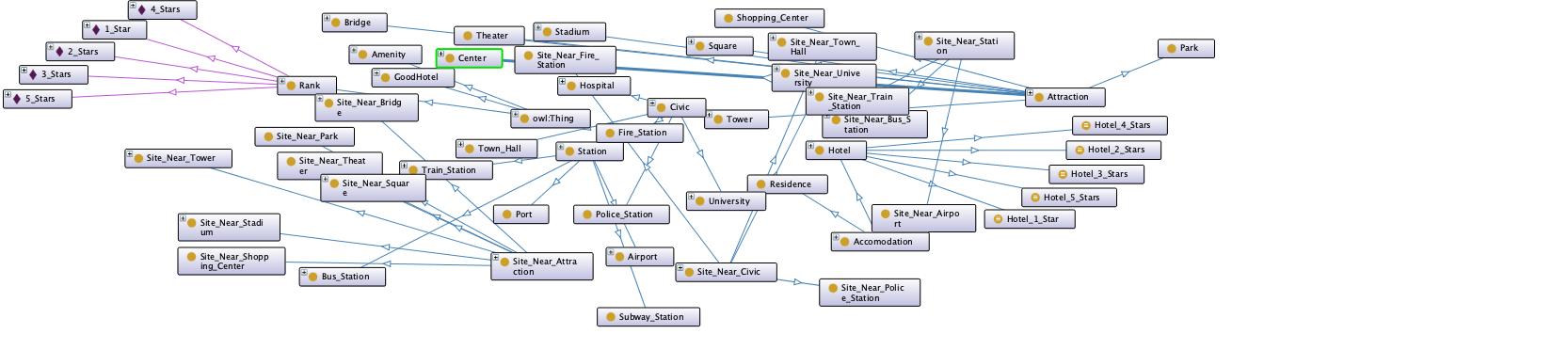}
\end{center}
  \caption{Excerpt of the Hotel ontology.}
  \label{hotelonto}
\end{figure}

Now, one may fix a city, say Pisa, extract the properties of the hotels from Web sites, such as location, price, etc.,  and
the  hotel judgements of the users, \eg, from Trip Advisor.\footnote{\url{http://www.tripadvisor.com}} Now,  using the terminology of the ontology and class instances gathered from the Web one may   ask about
what characterizes \emph{good} hotels in Pisa (our target class $T$) according to the user feedback. Then one may learn from the user feedback that, for instance, 
`an expensive hotel having as amenities babysitting, cradles, safety boxes and WI-FI is a good hotel to some degree $\m$'. 
\qed
\end{example}

\nd The objective is essentially the same as in \eg~\cite{Lisi15,Straccia15}, except that now we propose to rely on the $\mathbb{R}$eal AdaBoost~\cite{Nock07} boosting algorithm to be adapted to the (fuzzy) OWL case. Of course, like in~\cite{Lisi13a,Straccia15}, we continue to support so-called \emph{fuzzy concept descriptions} and \emph{fuzzy concrete domains} in the learned concept expressions~\cite{Lukasiewicz08a,Straccia05d,Straccia13} such as `an  \emph{expensive} Bed and Breakfast is a good hotel'. Here, the concept \emph{expensive} is a so-called fuzzy concept~\cite{Zadeh65}, \ie~a concept for which the belonging of an individual to the class is not necessarily a binary yes/no question, but rather a matter of degree in $[0,1]$. For instance, in our example, the degree of expensiveness of a hotel may depend on the price of the hotel: the higher the price the more expensive is the hotel.
Here, the range of the `attribute' \emph{hotel price} becomes a so-called \emph{fuzzy concrete domain}~\cite{Straccia13} allowing to specify fuzzy labels such as `high/moderate/low price'.

We recall that (discrete) AdaBoost~\cite{Freund96,Schapire99,Friedman00} uses weak hypotheses with outputs restricted to the discrete set of classes that it combines via leveraging weights in a linear vote. On the other hand $\mathbb{R}$eal AdaBoost~\cite{Nock07} is a generalisation of it as real-valued weak hypotheses are admitted (see~\cite{Nock07} for a comparison to approaches to real-valued AdaBoost).

Besides the fact that (to the best of our knowledge) the use of both (discrete) AdaBoost (with the notable exception of~\cite{Fanizzi19}) and its generalisation to real-valued weak hypotheses in the context OWL 2 ontologies is essentially unexplored, the main features of our algorithm, called \fuzzyowladaboost, are the following:
\begin{itemize}
\item it generates a set of  fuzzy $\el(\D)$ inclusion axioms~\cite{Bobillo18} that are the weak hypothesis, possibly including fuzzy concepts and fuzzy concrete domains \cite{Lukasiewicz08a,Straccia05d,Straccia13}. Each axiom has a leveraging weight;
\item the fuzzy concept inclusion axioms are then linearly combined into a new fuzzy concept inclusion axiom describing sufficient conditions for being an individual instance  of the target class $T$ and to which degree;
\item all generated fuzzy concept inclusion axioms could then be encoded as \emph{Fuzzy OWL 2} axioms~\cite{Bobillo10,Bobillo11c}. As a consequence, a Fuzzy OWL 2 reasoner, such as \emph{fuzzyDL}~\cite{Bobillo08a,Bobillo16}, can then be used to automatically determine (and to which degree) whether an individual belongs to the target class $T$.\footnote{Fuzzy OWL 2 and fuzzyDL need slightly to be extended to support the type of linear combination of weighted concepts we are going to use. The extension is straightforward.} 
\end{itemize}

\nd Let us remark that we rely on real-valued AdaBoost as the weak hypotheses \fuzzyowladaboost~generates are indeed fuzzy concept inclusion axioms and, thus, the degree to which an instance satisfies them is a real-valued degree of truth in $\unit$. 

In the following, we proceed as follows. In Section~\ref{sec:relatedWork} we compare our work with closely related work appeared so far. 
In Section~\ref{sec:background}, for the sake of completeness, we recap the salient notions we will rely on in this paper. Then, in Section~\ref{sec:learn} we will present our algorithm \fuzzyowladaboost~that then is evaluated for its effectiveness in Section~\ref{sec:eval}. Section~\ref{sec:conclusions} concludes and points to some topics of further research. 


\section{Related Work}
\label{sec:relatedWork}

\nd Concept inclusion axiom learning in DLs stems from statistical relational learning, where classification rules are (possibly weighted) Horn clause theories (see \eg~\cite{deRaedt08,DeRaedt17}), and various methods have been proposed in the DL context so far (see \eg~\cite{Lisi19,Rettinger12}). The general idea consists in the exploration of the search space of potential concept descriptions that cover the available training examples using so-called refinement operators (see, \eg~\cite{Badea00,Chitsaz12,Lehmann09,Lehmann07,Lehmann07a,Lehmann10,Lisi03}).
The  goal is then to learn a concept description of the underlying DL language covering (possibly) all the provided positive examples and (possibly) not covering any of the provided negative examples. The fuzzy case (see~\cite{Lisi13,Lisi15,Straccia15}) is a natural extension relying on fuzzy DLs~\cite{Bobillo15b,Straccia13} and fuzzy ILP (see \eg~\cite{SerrurierP07}) instead. 

Closely related to our work are~\cite{Fanizzi08,Fanizzi11,Fanizzi18,Fanizzi19,Lisi13,Lisi15,Straccia15}. In fact, \cite{Fanizzi08,Fanizzi11,Fanizzi18} are an adapation to the DL case of the the well-known  \foil-algorithm, while~\cite{Lisi13,Lisi15} that  stem essentially from~\cite{Lisi13a,Lisi14a,Lisi11a,Lisi11,Lisi13b,Lisi14}, propose \emph{fuzzy}  \foil-like algorithms instead, and are inspired by  fuzzy ILP variants such as~\cite{Drobics03,SerrurierP07,Shibata99}.\footnote{See, \eg~\cite{Cintra13}, for an overview on fuzzy rule learning methods.} Let us note that~\cite{Lisi13,Lisi11} consider the weaker hypothesis representation language \dllite~\cite{Artale09}, while  here we rely on a weighted sum of fuzzy $\el(\D)$ inclusion axioms, similarly to~\cite{Lisi13a,Lisi14a,Lisi11a,Lisi13b,Lisi14,Lisi15}. Fuzzy $\el(\D)$ has also been considered in~\cite{Straccia15}, which however differs from~\cite{Lisi13,Lisi15} by the fact that a (fuzzy) probabilistic ensemble evaluation of the fuzzy concept description candidates has been considered.\footnote{Also, to the best of our knowledge, concrete datatypes were not addressed in the evaluation.} Let us note that fuzzy $\el(\D)$  concept expressions are appealing as they can straightforwardly be translated into natural language and, thus, contribute to the explainability aspect of the induced classifier.

Discrete boosting has been considered in~\cite{Fanizzi19} that also shows how  to derive a weak learner (called \textsc{wDLF}) from conventional learners using some sort of random downward  refinement operator covering at least a positive example and yielding a minimal score fixed with a threshold. Besides that, we deal here with fuzziness in the hypothesis language and a real-valued variant of AdaBoost, the weak learner we propose here differentiates from the previous one by using a descent-like gradient algorithm to search for the best alternative. Notably, this  also deviates from `fuzzy' rule learning AdaBoost variants, such as~\cite{Jesus04,Otero06,Palacios11,Sanchez07,Zhu16} in which the weak learner is required to generate the whole rules' search space beforehand the selection of the best current alternative. 
Such an approach is essentially unfeasible in the OWL case due to the size of the search space.

\cite{Iglesias11} can learn fuzzy OWL DL concept equivalence axioms from FuzzyOWL 2 ontologies, by interfacing with the \emph{fuzzyDL} reasoner~\cite{Bobillo16}. The candidate concept expressions are provided by the underlying \textsc{DL-Learner}~\cite{Lehmann09a,Buehmann16,Buehmann18} system. However, it has been tested only on a toy ontology so far.
Last, but not least, let us mention~\cite{Konstantopoulos10} that is based on an ad-hoc translation of fuzzy \L{}ukasiewicz $\cal ALC$ DL constructs into fuzzy \emph{Logic Programming} (fuzzy LP) and uses a conventional  ILP method to learn rules. Unfortunately, the method is not sound as it has been shown that the mapping from fuzzy DLs to LP is incomplete~\cite{Motik07} and entailment in \L{}ukasiewicz $\cal ALC$ is undecidable \cite{Cerami13}. To be more precise, undecidability holds already for $\el$~under the infinitely valued \L{}ukasiewicz semantics~\cite{Borgwardt17a}.\footnote{We recall that $\el$ is a strict sub-logic of $\alc$.}

While it is not our aim here to provide an extensive overview about learning \wrt~ontologies literature, there are also alternative methods to what we present here.
So, \eg, the series of works \cite{Fanizzi10,Fanizzi10a,Rizzo14,Rizzo14a,Rizzo15,Rizzo15b,Rizzo16a,Rizzo17,Rizzo18} are inspired on \emph{Decision Trees/Random Forests},
 \cite{Bloehdorn07,Fanizzi06,Fanizzi08a,Fanizzi12a} consider \emph{Kernel Methods} for inducing concept descriptions, while \cite{Minervini11,Minervini12,Minervini12a,Minervini14,Zhu15}  consider essentially a \emph{Naive Bayes} approach. Last but not least, \cite{Lehmann07b} is inspired on \emph{Genetic Programming} to induce concept expressions, while \cite{Nickles14} is based on the \emph{Reinforcement Learning} framework.
 



\section{Background}
\label{sec:background}

\nd For the sake of completeness, we recap here the salient notions about \emph{fuzzy Desciption Logics} (fuzzy DLs) we will rely on in this paper. 
The interested reader may refer to \eg~\cite{Bobillo15b,Straccia13} for a more in depth description of the various notions introduced here.


%
\begin{figure}
\begin{center}
\begin{tabular}{c@{\ \ \ \ \ \ \ \ }c@{\ \ \ \ \ \ \ \ }c@{\ \ \ \ \ \ \ \ }c}
\includegraphics[scale=0.4]{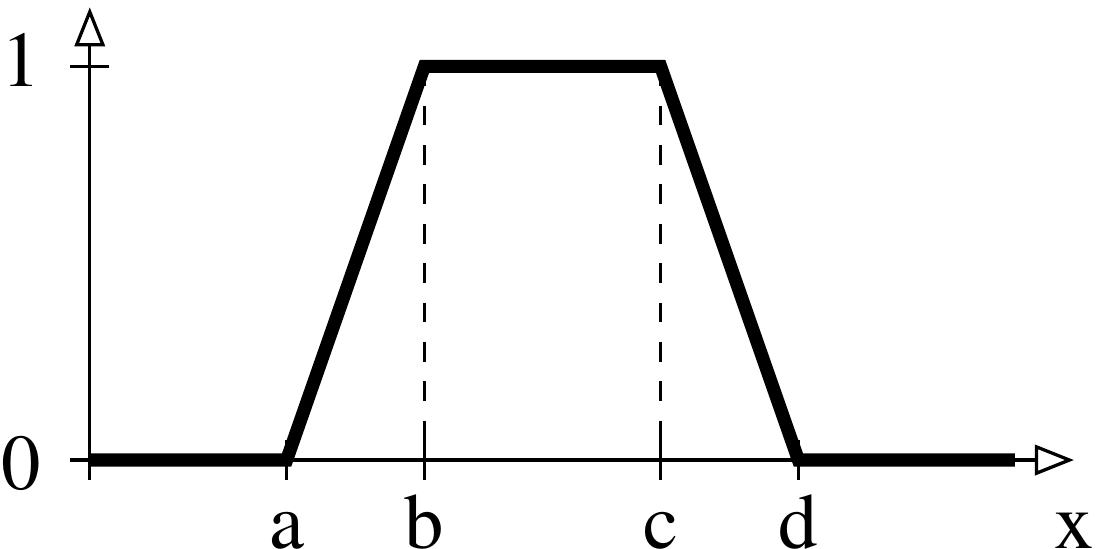} &
\includegraphics[scale=0.4]{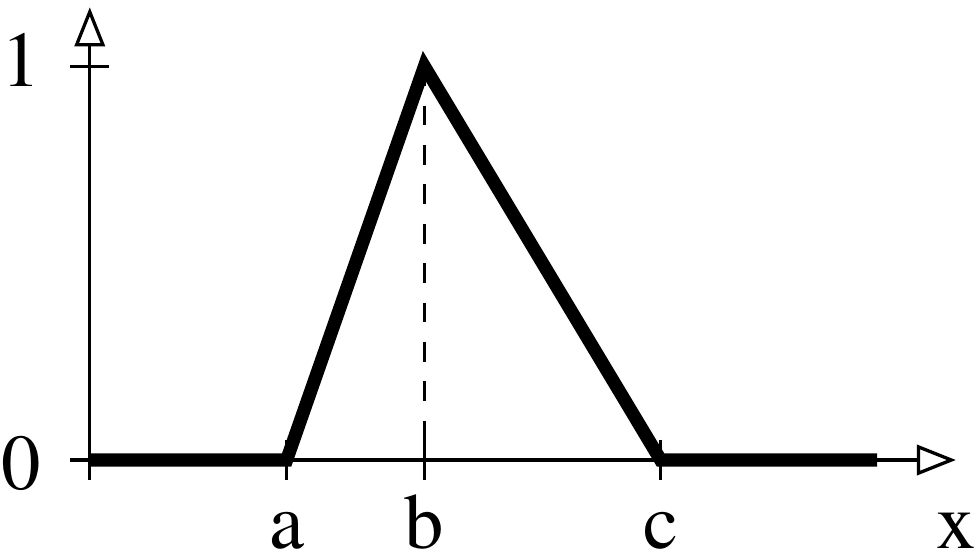} & \\
(a) & (b)   \\
\includegraphics[scale=0.4]{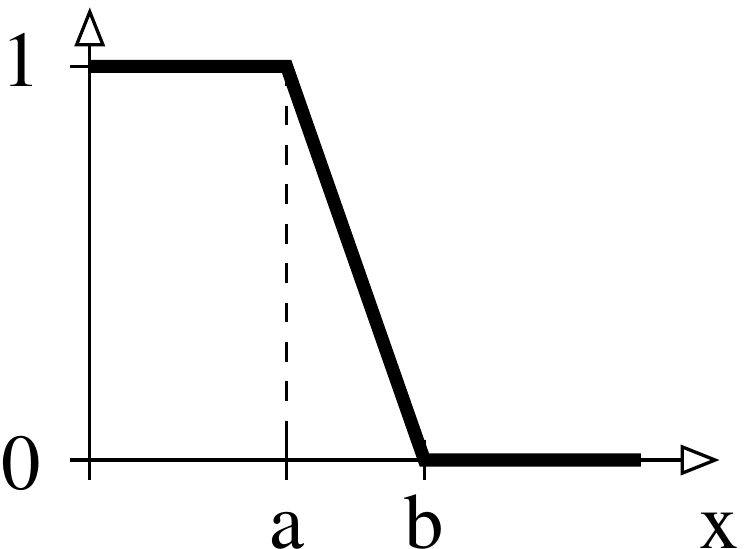} &
\includegraphics[scale=0.4]{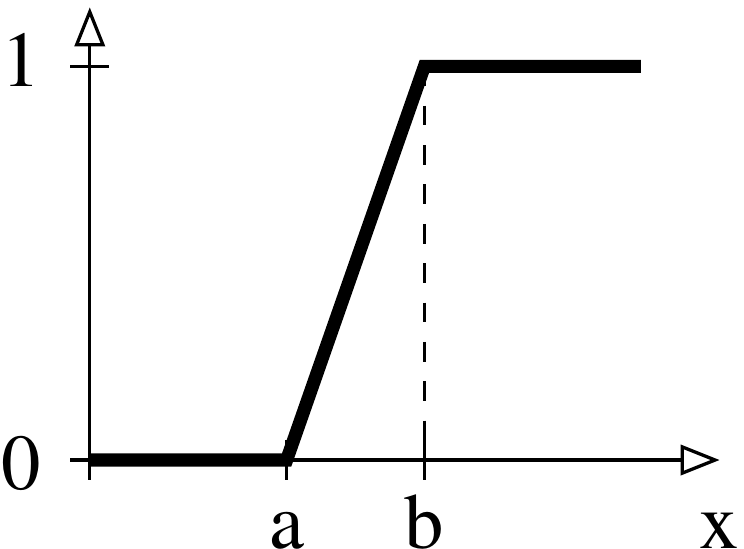} \\
(c) & (d)
\end{tabular}
\caption{(a) Trapezoidal function $\mathit{trz}(a,b,c,d)$,
(b) triangular function $\mathit{tri}(a,b,c)$, (c) left shoulder
function~$\mathit{ls}(a,b)$, and (d) right shoulder function $\mathit{rs}(a,b)$.}
\label{fig:muf}
\end{center}
\end{figure}
%
%
%
\paragraph{Fuzzy Sets}
To start with, we recap that a \emph{fuzzy set} $A$ over
a countable crisp set $X$ is a function $A \colon X \to [0,1]$, called \emph{fuzzy membership} function of $A$~\cite{Zadeh65}. A crisp set $A$ is defined by  a membership function   $A \colon X \to \{0,1\}$ instead. The `standard' fuzzy set operations conform to 
$(A \cap B)(x) = \min(A(x), B(x))$, $(A \cup B)(x) = \max(A(x), B(x))$ and $\bar{A}(x) = 1- A(x)$ ($\bar{A}$ is the set complement of $A$), 
the \emph{cardinality} of a fuzzy set is defined as  
\begin{equation} \label{sigma}
|A| = \sum_{x\in X} A(x) \ , 
\end{equation}

\nd while  the \emph{inclusion degree} of $A$ in $B$ is defined as 

\begin{equation} \label{incdeg}
ideg(A,B) = \frac{|A \cap B|}{|A|} \ . 
\end{equation}

\nd The trapezoidal (Fig.~\ref{fig:muf} (a)), the triangular
(Fig.~\ref{fig:muf} (b)), the $L$-function (left-shoulder function,
Fig.~\ref{fig:muf} (c)), and the $R$-function (right-shoulder
function, Fig.~\ref{fig:muf} (d)) are frequently used to specify
membership functions of fuzzy sets.

\begin{figure}
\begin{center}
\begin{tabular}{cc}
\includegraphics[scale=0.5]{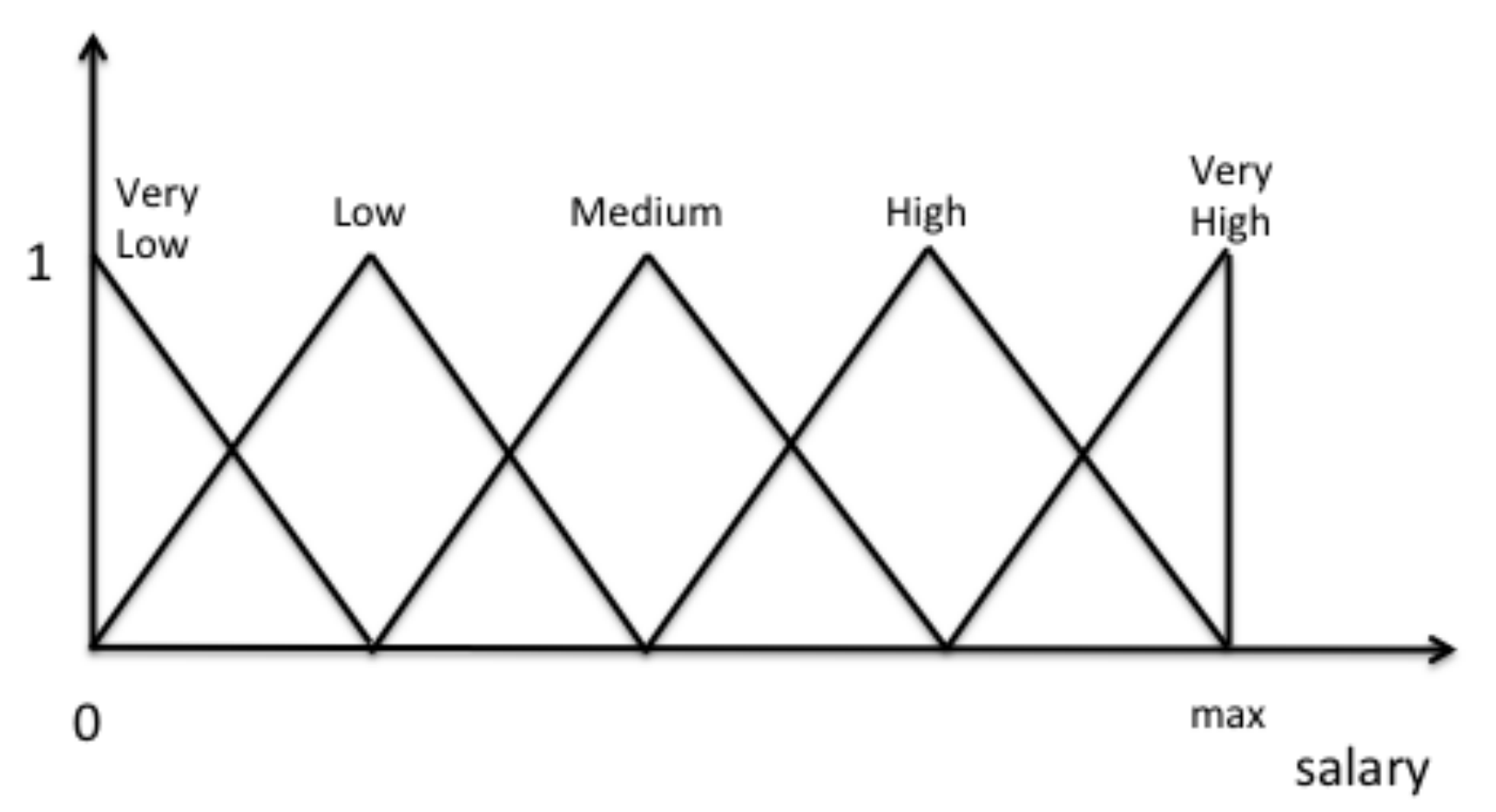}
\end{tabular}
\caption{Uniform fuzzy sets over salaries.}
\label{partfuzzytrz}
\end{center}
\end{figure}

Although fuzzy sets have a  greater expressive power than
classical crisp sets, their usefulness depends critically on the
capability to construct appropriate membership functions for various
given concepts in different contexts.  We refer the
interested reader to, \eg,~\cite{Klir95}. 
One easy and typically satisfactory method to define the membership
functions is to uniformly partition the range of, \eg~salary values
(bounded by a minimum and maximum value), into 3, 5 or 7 fuzzy sets
using triangular (or trapezoidal) functions (see Fig.~\ref{partfuzzytrz}). Another popular approach may consist in
using the so-called \emph{C-means} fuzzy clustering algorithm (see, \eg~\cite{Bezdek81}) with 3,5 or or 7 clusters, 
where the fuzzy membership functions are triangular functions built around the centroids of the clusters (see \eg~also~\cite{Huitzil20,Huitzil18}).

%

\paragraph{Fuzzy Description Logics}
Next, we recap here the fuzzy DL~$\ELNW(\D)$ extending the well-known  fuzzy DL $\EL$~\cite{Bobillo15b,Straccia13} with\footnote{For classical $\EL$ and DLs in general we refer the reader to~\cite{Baader05a,Baader07a}.} 
\begin{itemize}

\item  \emph{real-valued weighted concept} construct (denoted by the letter $\mathcal{W}$)~\cite{Bobillo11c,Straccia13};\
\item \emph{atomic negation} (denoted by  $\neg$); 
\item fuzzy concrete domains (denoted by $\D$)~\cite{Straccia05d}. 
\end{itemize}

\nd Fuzzy $\ELNW(\D)$ is expressive enough to capture all of the ingredients we are going to use in this work. Of course, DLs, fuzzy DLs, OWL 2 and fuzzy OWL 2 in particular,  cover many more language constructs than we use here (see, \eg~\cite{Baader07a,Bobillo15b,Bobillo11c,Straccia13}).

From a syntax point of view, we start with the notion of \emph{fuzzy concrete domain}, that is a tuple
$\D\eqs\tuple{\Delta^{\D}, {\,\cdot\,}^{\D}}$ with datatype
domain~$\Delta^{\D}$ and a mapping ${\,\cdot\,}^{\D}$ that assigns
to each data value an element of $\Delta^{\D}$, and to every $1$-ary
datatype predicate $\mathbf{d}$  (see below) a $1$-ary fuzzy relation over $\Delta^{\D}$. 
Therefore, ${\,\cdot\,}^{\D}$  maps indeed each datatype predicate $\mathbf{d}$ 
into a function from $\Delta^{\D}$ to $[0,1]$. The domains we consider here are the integers, the reals and the booleans. The datatype predicates are defined as 
%
%
\begin{eqnarray*}
\mathbf{d} & \rightarrow &  ls(a,b)  \ | \ rs(a,b) \ | \   tri(a,b,c)  \ | \ trz(a,b,c,d)  \ | \  \ = _{v}  \ ,
\end{eqnarray*}
\nd where $v$ is an integer, real or boolean value, \eg~$ls(a,b)$ is the left-shoulder membership function (see Fig.~\ref{fig:muf}) and $= _{v}$ corresponds to the crisp singleton set $\{v\}$.

%

Now, consider pairwise disjoint alphabets ${\bf I}, {\bf A}$ and ${\bf R}$, 
where  ${\bf I}$ is the set of \emph{individuals}, 
${\bf A}$ is the set of \emph{concept names} (also called \emph{atomic concepts})
and ${\bf R}$ is the set of \emph{role names}. 
Each role is either an  \emph{object property} or  a \emph{datatype property}.
The set of fuzzy $\ELNW(\D)$ \emph{concepts} are built from concept names $A$  using
connectives and quantification constructs over object properties $r$
and  datatype properties $s$, 
as described by the following syntactic rule ($n\geq 1, \alpha_i \in \mathbb{R} \setminus \{0\}$):
%
\begin{eqnarray*}
C  & \rightarrow &  \topc \ | \ \bottomc \ | \ A\ |\  \notc A \ | \ C_{1} \andc C_{2} \ |\  \some r.C \ | \ \ |\   \some s.\mathbf{d}  \ |\  \alpha_1 \cdot A_1 + \ldots + \alpha_n \cdot A_n \ .
\end{eqnarray*}

\nd Note that we generalise slightly the notion of weighted sum of Fuzzy OWL 2 in which $\alpha_i \in (0,1], \sum_i \alpha_i \leq 1$ is assumed. 

%
%

A \emph{fuzzy assertion} axiom is an expression of the form
\fuzzyg{\cass{a}{C}}{\m} (called \emph{fuzzy concept assertion},  -- $a$ is an
instance of concept $C$ to degree greater than or equal to $\m$) or of the form
$\fuzzyg{\rass{a_{1}}{a_{2}}{r}}{\m}$ (called \emph{fuzzy role assertion},
-- $(a_{1}, a_{2})$ is an instance of object property $r$ to degree greater than or equal to
$\m$), where $a, a_{1}, a_{2}$ are individual names, $C$ is a
concept, $r$ is an object property and $\m \in \unitp$.

A \emph{fuzzy General Concept Inclusion} (fuzzy GCI) axiom  is of the form
$\fuzzyg{C_{1} \impc C_{2}}{\m}$ ($C_{1}$ is a sub-concept of $C_{2}$ to degree greater than or equal to $\m$), where $C_{i}$ is a concept and 
$\m \in \unitp$. We may also call a fuzzy GCI of the form $\fuzzyg{C \impc A}{d}$, where $A$ is a concept name, a \emph{rule}, where $A$ is called the \emph{head} and $C$ is called the \emph{body} of the rule.


For ease of presentation, in case  the degree $\m$ is $1$, we simply omit the degree $1$ and write, \eg~$\cass{a}{C}$ (resp.~$C_{1} \impc C_{2}$) in place of \fuzzyg{\cass{a}{C}}{1} (resp.~$\fuzzyg{C_{1} \impc C_{2}}{1}$).  We also write $C_1 = C_2$ as a macro for the two GCIs $C_1 \impc C_2$ and $C_2 \impc C_1$, indicating that the two concepts $C_1$ and $C_2$ are equivalent.

\emph{Fuzzy $\el(\D)$} is the language fuzzy $\ELNW(\D)$ without atomic negation and weighted sum, 
while \emph{crisp $\el(\D)$}, or simply  $\el(\D)$, is fuzzy  $\el(\D)$ without $ls, rs, tri$ and $trz$ datatype  predicates and the degree in axioms is restricted to be $1$.


A (crisp) \emph{Knowledge Base} (KB) $\K$ is a set of (crisp) $\el(\D)$  assertions and GCIs, while  a \emph{fuzzy Knowledge Base} (fKB) $\fKB$ is a set of fuzzy $\ELNW(\D)$  assertions and GCIs.
\begin{remark} \label{elkb}
We anticipate that in our setting we are going to learn fuzzy GCIs from \emph{crisp} OWL 2 data, \ie~we are going to learn from crisp knowledge bases only.
\end{remark}
With $\indkb$ we denote the set of individuals occurring  in a KB  $\KB$.

\begin{example}[Example~\ref{runes} cont.] \label{ex:hotel}
Related to Example~\ref{runes}, an example of GCI is 
\[
\mathsf{Hotel} \impc \mathsf{Accommodation}
\]
\nd (a hotel is an accommodation), while the assertions  
\[
\begin{array}{l}
\rass{\mathsf{Hotel\_010}}{\mathsf{3\_Stars}}{\mathsf{hasRank}} \\
\cass{\mathsf{Hotel\_010}}{\mathsf{Hotel} \andc (\csome \mathsf{hasAmenity}.\mathsf{24h\_Reception}) \andc (\csome \mathsf{hasPrice}.=_{\mathsf{79}})}
\end{array}
\]

\nd describe a three star hotel ($\mathsf{Hotel\_010}$) that provides 24h reception as amenity and whose room price is 79 (euro).
\qed
\end{example}

\nd Concerning the semantics, let us fix a fuzzy concrete domain 
$\D\eqs\tuple{\Delta^{\D}, {\,\cdot\,}^{\D}}$ over integer, reals and booleans, with fuzzy membership functions $ls, rs, tri$ and $trz$ over integers and reals, and equality predicate $=_{\cdot}$ over integers, reals and booleans.

Now, unlike classical DLs in which an interpretation $\I$
maps \eg~a concept $C$ into a set of individuals $\highi{C}
\subseteq \highi{\Delta}$,  \ie~$\I$ maps $C$ into a function
$\highi{C}: \highi{\Delta} \to \{0, 1\}$ (either an individual
belongs to the extension of $C$ or does not belong to it), in fuzzy
DLs, $\I$ maps $C$ into a function $\highi{C}: \highi{\Delta} \to
\unit$ and, thus, an individual belongs to the extension of $C$ to
some degree in $[0,1]$, \ie~$\highi{C}$ is a fuzzy set.

Specifically, a \emph{fuzzy interpretation} is a pair $\I =
(\highi{\Delta}, \highi{\cdot})$ consisting of a nonempty (crisp)
set $\highi {\Delta}$ (the \emph{domain}) and of a \emph{fuzzy
interpretation function\/} $\highi{\cdot}$ that assigns: \ii{i} to
each atomic concept $A$ a function $\highi{A}\colon\highi{\Delta}
\rightarrow \unit$; \ii{ii} to each object property $r$ a function
$\highi{r}\colon\highi{\Delta} \times \highi{\Delta} \rightarrow
\unit$; \ii{iii} to each datatype property $s$ a function
$\highi{s}\colon\highi{\Delta} \times \Delta^{\D} \rightarrow
\unit$; \ii{iv} to each individual $a$ an element $\highi{a} \in
\highi{\Delta}$ such that $\highi{a} \neq \highi{b}$ if $a \neq b$ (the so-called \emph{Unique Name Assumption}); and \ii{v} to each data value $v$ an element
$\highi{v} \in \Delta^{\D}$.
Now, a fuzzy interpretation function is extended to 
concepts via standard fuzzy set operations as specified below 
(where $x \in \highi{\Delta}$):\footnote{The semantics of the weighted sum will be clearer once we address the learning problem.}
%

\begin{eqnarray*}
\highi{\topc}(x)  & =  & 1 \\
\highi{\bottomc}(x)  & =  & 0  \\
\highi{(C \andc D)}(x)  & =  & \min(\highi{C}(x), \highi{D}(x))  \\
\highi{(\notc A)}(x)  & =  & 1- \highi{A}(x) \\
\highi{(\csome r.C)}(x)  & =  &  \sup_{y \in \highi{\Delta}} \{ \min(\highi{r}(x,y), \highi{C}(y)) \} \\
\highi{(\csome s.\mathbf{d})}(x)  & =  &  \sup_{y \in  \Delta^{\D}} \{ \min( \highi{s}(x,y), \mathbf{d}^{\D}(y)) \} \\
\highi{(\alpha_1 \cdot A_1 + \ldots + \alpha_n \cdot A_n)}(x)  & =  & \left \{
\begin{array}{ll}
1 & \mbox{if } \sum_i \alpha_i \cdot \highi{A_i}(x) > 1 \\
0 & \mbox{if } \sum_i \alpha_i \cdot \highi{A_i}(x) < 0 \\
\sum_i \alpha_i \cdot \highi{A_i}(x)  & \text{else}  \ .
\end{array} \right . 
\end{eqnarray*}

%
%
%
\nd The \emph{satisfiability of axioms} is  defined by the
following conditions:
\ii{i} $\I$ satisfies $\fuzzyg{\cass{a}{C}}{\m}$ if    $C^{\I} (a^{\I}) \geq \m$;
\ii{ii} $\I$ satisfies $\fuzzyg{\rass{a}{b}{r}}{\m}$ if  $r^{\I} (a^{\I}, b^{\I}) \geq \m$; and
 \ii{iii} $\I$ satisfies $\fuzzyg{C \impc D}{\m}$ if  for all $x \in  \highi{\Delta}$, $\highi{D}(x) \geq  \highi{C}(x) \cdot \m$.

Now, consider a  set $\calS$ of axioms. Then, \ii{i} $\I$ is  a \emph{model} of  $\calS$ if $\I$ satisfies each axiom in $\calS$; 
\ii{ii} $\calS$ is \emph{satisfiable} (or \emph{consistent}) if $\calS$ has a model;
\ii{iii} $\calS$ \emph{entails}  axiom $\phi$, denoted $\calS \models \phi$, if every model of $\calS$ satisfies
$\phi$; and \ii{iv} the \emph{best entailment degree} of $\phi$ of the form $C \impc D$, $\cass{a}{C}$ or $\rass{a}{b}{r}$, denoted $\bed{\calS}{\phi}$, is defined as
\[
\bed{\calS}{\phi} = \sup\{ \m \mid \calS \models \fuzzyg{\phi}{\m}\} \ .
\]


%

%
\begin{example}[Example~\ref{ex:hotel} cont.] \label{ex:hotelB}
%

\nd Consider a KB $\KB$ whose excerpt is described in Example~\ref{ex:hotel}. Consider the following fuzzy GCI $\phi$
\[
\fuzzyg{\mathsf{Accommodation}  \andc (\some \mathsf{hasPrice}.\mathsf{Fair}) \impc \mathsf{GoodHotel}}{0.56}\ ,
\]
\nd where $\mathsf{hasPrice}$ is a datatype property whose values are
measured in euros and the price concrete domain has been
fuzzified as illustrated in Fig.~\ref{fig:hasPrice-fuzzy}. 
The intended meaning of this axiom is roughly  `an accommodation with a fair room price is to some degree a good hotel'.
Also consider $\mathsf{Hotel\_010}$ in Example~\ref{ex:hotel}.

\begin{figure}
\begin{center}
\includegraphics[scale=0.5]{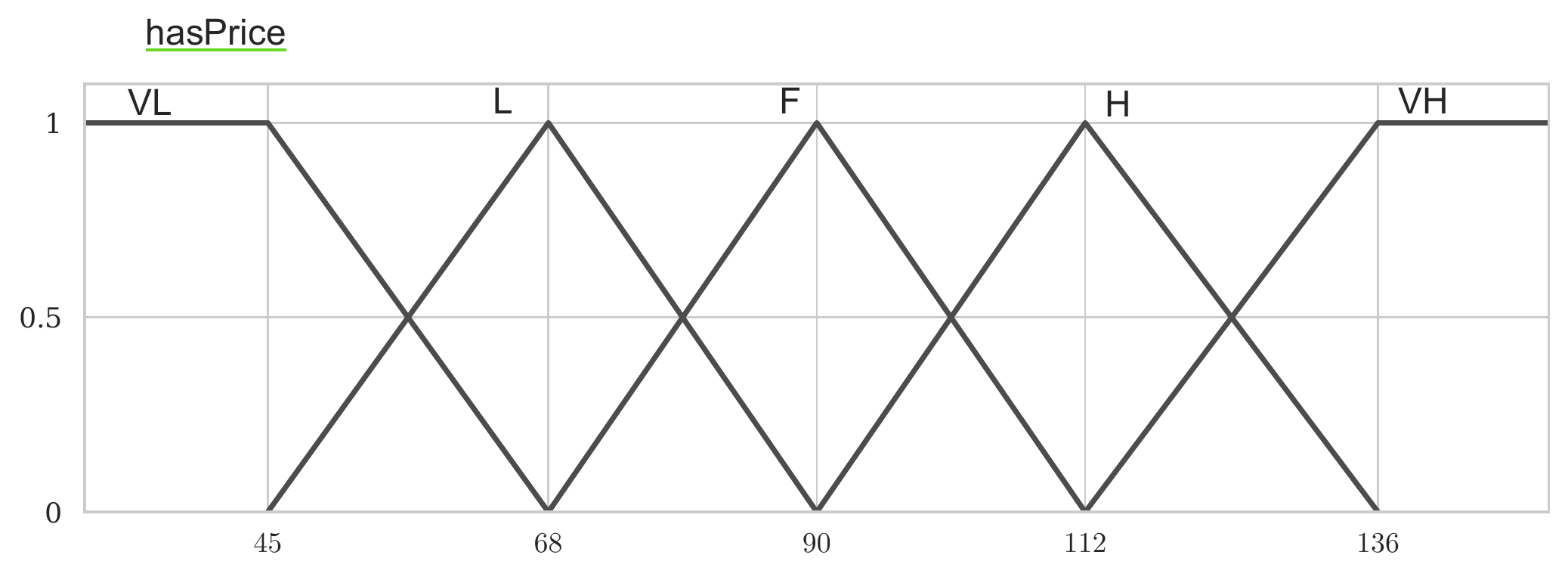}
\end{center}
  \caption{Fuzzy sets derived from the datatype property $\mathtt{hasPrice}$.}
  \label{fig:hasPrice-fuzzy}
\end{figure}

Now, it can be verified that
\[
\KB \cup \{\phi\} \models \fuzzyg{\cass{\mathsf{Hotel\_010}}{\mathsf{GoodHotel}}}{0.28} \ ,
\]
\nd and, more specifically that 
\[
\bed{\KB \cup \{\phi\}}{\cass{\mathsf{Hotel\_010}}{\mathsf{GoodHotel}}} =0.28 \ ,
\]
\nd which indicates to which degree $\mathsf{Hotel\_010}$ is a $\mathsf{GoodHotel}$.

\qed
\end{example}

\nd Finally, consider a fuzzy concept $C$, a fuzzy GCI $C \impc D$, a KB $\KB$, a set of individuals $\mathsf{I}$ and a (weight) distribution $\myvec{w}$ over $\ind$. Then
the \emph{cardinality} of $C$ \wrt~$\KB$ and $\mathsf{I}$, denoted $|C|_\KB^{\mathsf{I}}$, is defined as (\cf~Eq.~\ref{sigma})
\begin{equation} \label{card}
|C|_\KB^{\ind} = \sum_{a \in \ind} \bed{\KB}{\cass{a}{C}} \ ,
\end{equation}
\nd while the \emph{weighted cardinality} $C$ \wrt~$\KB$, $\myvec{w}$ and $\ind$ , denoted $|C|_\KB^{\myvec{w},\ind}$, is defined as\footnote{The weight of $a \in \ind$ \wrt~$\myvec{w}$ is denoted as $w_a$.} 
\begin{equation} \label{wcard}
|C|_\KB^{\myvec{w},\ind} = \sum_{a \in \ind} w_a \cdot \bed{\KB}{\cass{a}{C}} \ .
\end{equation} 

\nd The \emph{crisp cardinality} (denoted $\ceil{C}_\KB^{\ind}$) and \emph{crisp weighted cardinality}  (denoted $\ceil{C}_\KB^{\myvec{w},\ind}$) are defined similarly by replacing in Eq.~\ref{card} and \ref{wcard} the term $\bed{\KB}{\cass{a}{C}}$ with $\ceil{\bed{\KB}{\cass{a}{C}}}$.

Furthermore, the \emph{confidence degree} (also called  \emph{inclusion degree}) of $C \impc D$ \wrt~$\KB$ and $\mathsf{I}$, denoted $cf(C \impc D, \ind)$, is defined as (\cf~Eq.~\ref{incdeg})
\begin{equation}\label{cf}
cf(C \impc D, \ind) =  \frac{|C \andc D |_\KB^{\ind}}{|C|_\KB^{\ind}} \ .
\end{equation}

\nd Similarly,  the \emph{weighted confidence degree} (also called  \emph{weighted inclusion degree}) of $C \impc D$ \wrt~$\KB$, $\myvec{w}$ and $\mathsf{I}$, denoted $cf(C \impc D, \myvec{w}, \ind)$, is defined as
\begin{equation}\label{wcf}
cf(C \impc D, \myvec{w},\ind) =  \frac{|C \andc D |_\KB^{\myvec{w},\ind}}{|C|_\KB^{\myvec{w},\ind}} \ .
\end{equation}

\section{Learning Fuzzy Concept Inclusions via Real-Valued Boosting}
\label{sec:learn}

\nd To start with, we introduce our learning problem.

\subsection{The Learning Problem} \label{sect:tlp}
In general terms, the learning problem we are going to address is stated as follows: 

{\bf Given:}
\begin{itemize}

\item a satisfiable (crisp) KB $\K$ and its individuals $\indkb$;

\item a \emph{target concept name} $T$ with an associated unknown  classification function 
$f_T \colon \indkb \to \{1,0\}$,
where for each $a\in \indkb$, the possible values (\emph{labels}) correspond, respectively, to  $\K \models \cass{a}{T}$ ($a$ is a \emph{positive} example of $T$) and $\K \not \models \cass{a}{T}$ ($a$ is a \emph{non-positive} example of $T$);

\item a \emph{hypothesis space} of classifiers $\mathcal{H} = \{h \colon \indkb \to [0,1]\}$;

\item a \emph{training set} $\mathcal{E} = \mathcal{E}^{+} \cup \mathcal{E}^{-}$ (the positive and non-positive examples of $T$, respectively) of individual-label pairs:
\begin{eqnarray*}
\mathcal{E}^{+} & = & \{(a,1) \mid a \in \indkb, f_T(a) = 1 \} \\
\mathcal{E}^{-} & = & \{(a,0) \mid a \in \indkb, f_T(a) = 0 \}  \ .
\end{eqnarray*}
\nd With $\inds{\mathcal{E}}$ we denote the set of individuals occurring in $\mathcal{E}$. We assume that for all $a\in \inds{\mathcal{E}}$, $0 = \bed{\KB}{\cass{a}{T}} = \bed{\KB}{\cass{a}{\neg T}}$. That is we state that $\K$ does not already know whether $a$ is an instance of $T$ or of $\notc T$. We write $\E(a) =  1$ if $a$ is a positive example (\ie, $a \in  \inds{\E^+}$), $\E(a) =  0$ if $a$ is a non-positive example (\ie, $a \in  \inds{\E^-}$).
\end{itemize}

\nd {\bf Learn:} a classifier $\bar{h} \in \mathcal{H}$ that is the result of \emph{Emprical Risk Minimisation} (ERM) on $\mathcal{E}$. That is, 
\begin{eqnarray*}
\bar{h} & = & \arg \min_{h \in \mathcal{H}} R(h, \mathcal{E}) \\
& = &  \mathbf{E}_{\mathcal{E}}[L(h(a),\E(a))] \\
& = & \frac{1}{|\mathcal{E}|} \sum_{a \in \inds{\E}} L(h(a),\E(a)) \ ,
\end{eqnarray*}

\nd where $L$ is a \emph{loss function} such that $L(\hat{l}, l)$  measures how different the prediction $\hat{l}$ of a hypothesis is from the true outcome $l$ and $R(h,\mathcal{E})$ is the \emph{risk} associated with hypothesis $h$ over $\mathcal{E}$, defined as the expectation of the loss function over $\mathcal{E}$. 

The effectiveness of the learned classifier $\bar{h}$ is then assessed by determining $R(\bar{h}, \mathcal{E}') $ on a a \emph{test set} $\mathcal{E}'$, disjoint from $\mathcal{E}$. 

In our learning setting, we assume that a hypothesis $h \in \mathcal{H}$ is a  set of fuzzy GCIs that has the form
\begin{equation} \label{eqnhyp}
\begin{array}{rcl}
\alpha_1 \cdot WL_1 + \ldots + \alpha_n \cdot WL_n &  \impc  & T \\
C_{ij} & \impc  & WL_i  \ , \text{ with }1 \leq i \leq n, 1 \leq j \leq k_i \ ,
\end{array}
\end{equation}

\nd where each $WL_i$ is a new atomic symbol not occurring in the KB and were each $C_{ij}$ is a fuzzy 
$\elbool$
concept expression defined as (where $v$ is a boolean value)
\begin{equation*}\label{eq:left-hand-GCI}
\begin{array}{lcl}
C & \longrightarrow &  \top \mid A \mid \some r.C \mid \some s.\mathbf{d} \mid C_{1} \andc C_{2} \\
\dt & \rightarrow  & ls(a,b) \ | \  rs(a,b) \ | \ tri(a,b,c) \  | \  trz(a,b,c,d) \ | \  = _{v} \ .
\end{array}
\end{equation*}

\nd  Essentially, each $\alpha_i$ indicates how well the `union' of the $C_{i1}, \ldots, C_{ik_i}$ (the \emph{Weak Learner} $WL_i$) contributes to classify an individual $a$ as being an instance of $T$. Specifically, if $\alpha_i > 0$ then $WL_i$ contributes to $a$'s positiveness, while if $\alpha_i < 0$ then $WL_i$ contributes to $a$'s non-positiveness instead.


\begin{remark}\label{el-}
Please note that we do not learn expressions of the form \eg~$\csome s.= _{v}$ for integer/real values $v$ as the search space would be too large and they would be likely non-effective. This is the reason why we restrict the $C_{ij}$ in Eq.~\ref{eqnhyp} to fuzzy $\elbool$ concept expressions and not fuzzy $\el(\D)$ instead.
\end{remark}

\nd For  $a\in \indkb$, the \emph{classification prediction value} $h(a)$ of $a$ \wrt~$h$, $T$ and $\K$ is defined as (for ease, we omit $\K$ and $T$)
\[
h(a) = \bed{\KB \cup h }{\cass{a}{T}} \ .
\]

\nd Note that, as stated above, essentially a hypothesis is a sufficient condition (expressed via the weighted sum of concepts) for being an individual instance of a target concept to  some degree. If $h(a) = 0$ then we say that $a$ is a non-positive instance of $T$, while if $h(a) > 0$ then $a$ is a positive instance of $T$ to  degree $h(a)$. 


\begin{remark}
Clearly,  the set of hypothesis by this syntax is potentially infinite due, \eg, to conjunction and the nesting of existential restrictions in the $C_{ij}$. This set is made finite by imposing further restrictions on the generation process such as the maximal number of conjuncts and the maximal depth of existential nestings allowed.\qed
\end{remark}

\begin{remark} \label{negativeex}
One may also think of further partitioning the set $\calE^-$ of non-positive examples into a set $\calE^\neg$ of \emph{negative} and a set $\calE^u$ of \emph{unknown} examples (and use as labelling set $\{-1,0,1\}$, respectively, with $1$ --positive, $0$ -- unknown, $-1$ -- negative), as done in some other approach (see \eg~\cite{Fanizzi19}). That is,  an individual $a$ is a \emph{negative} example of $T$ if $\K \models \cass{a}{\neg T}$, while  $a$ is an \emph{unknown} example of $T$ if neither $\K \models \cass{a}{T}$ nor $\K \models \cass{a}{\neg T}$ hold. In that case, usually we are looking for an exact definition of $T$, \ie~a hypothesis is of the stronger form $T = C$ instead.\footnote{We recall that a hypothesis as in Eq.~\ref{eqnhyp} does not allow us to infer negative instances of $T$, while $T=C$ does.}  
Which one to choose may depend on the application domain and on the effectiveness of the approach.~We do not address this case here.
\qed
\end{remark}

\nd We conclude with the notions of \emph{consistent}, \emph{non-reduntant},  \emph{sound}, \emph{complete} and \emph{strongly complete} hypothesis $h$  \wrt~$\K$, which are defined as follows:
\begin{description}
	\item[Consistency.] $\mathcal{K} \cup h$ is a consistent;
	\item[Non-Redundancy.] $\mathcal{K} \not\models \phi$, for all $\phi \in h$.
	\item[Soundness.]  $\forall  a \in \inds{\mathcal{E}^{-}}, h(a) = 0$.
	\item[Completeness.]  $\forall  a \in \inds{\mathcal{E}^{+}}, h(a) > 0$.
	\item[Strong Completeness.]  $\forall  a \in \inds{\mathcal{E}^{+}}, h(a) =1$.
\end{description} 

\nd We say that a hypothesis $h$ \emph{covers} (\emph{strongly covers}) an example $e \in \mathcal{E}$ iff $\bed{\KB \cup h}{e} > 0$ ($\bed{\KB \cup h}{e} = 1$). Therefore, soundness states that a  learned hypothesis is not allowed to cover a non-positive example, while the way (strong) completeness is stated  guarantees that all positive examples are (strongly) covered.

In general a learned (\emph{induced}) hypothesis $h$ has to be \emph{consistent}, \emph{non-reduntant} and  \emph{sound} \wrt~$\K$, but not necessarily complete, but, of course, these conditions can also be relaxed.

\subsection{The Learning Algorithm \fuzzyowladaboost} \label{sect:tlpa}

\nd We now present our real-valued boosting-based algorithm, which is based on a boosting schema (this section) applied to a fuzzy $\elbool$  \emph{weak learner} described in more detail in Section~\ref{sect:wla}.
Our learning method creates an \emph{ensemble} of fuzzy GCIs (see Eq.~\ref{eqnhyp}):
essentially, at each iteration our boosting algorithm invokes a weak learner that generates a  set of fuzzy $\elbool$ candidate GCIs that has the form $h_i = \{ C_{i1} \impc T, \ldots, C_{ik_i} \impc T \}$, called \emph{weak hypothesis},  determining  a change to the distribution of the weights associated with the examples.  The weights of misclassified examples get increased so that a better classifier can be produced in the next round, indicating the harder examples to focus on. The weak hypotheses are then  combined into a final hypothesis via a weighted sum of the weak hypotheses. 
We will rely on $\mathbb{R}$eal AdaBoost~\cite{Nock06,Nock07} as boosting algorithm, while we will use a weak learner that is similar to  \foildl~\cite{Lisi13,Lisi13a,Lisi15}, both of which need to be adapted to our specific setting.

Formally, consider a KB $\K$, a training set $\mathcal{E}$, a set of individuals $\ind$ with $\inds{\E} \subseteq \ind \subseteq \indkb$, and a weight distribution $\myvec{w}$ over $\ind$. With $\myvec{u}$ we indicate the uniform distribution over $\ind$, \ie~$u_a = 1/|\ind|$ (with $a \in \ind$). Furthermore, consider a  \emph{weak hypothesis} $h_i$, \ie~a set
$h_i = \{ C_{i1} \impc T, \ldots, C_{ik_i} \impc T \}$ of fuzzy $\elbool$ GCIs returned by the weak learner. Note that for
$a \in \indkb$, $\bed{\K\cup h_i}{\cass{a}{T}} \in[0,1]$. Next, we transform this value into a value in $[-1,1]$ as required by $\mathbb{R}$eal AdaBoost. So, let
$t\colon [0,1] \to [-1,1]$ be the transformation function 
\begin{equation*}
t(x) = 
\begin{cases*}
-1 & if $x =0$ \\
x & else 
\end{cases*}
\end{equation*}
\nd and let the classification prediction value $h_i(a)$ of $a$ \wrt~$h$, $T$ and $\K$ be  defined as (again for ease, we omit $\K$ and $T$)
\[
h_i(a) = t(\bed{\KB \cup h_i }{\cass{a}{T}}) \in \{-1\} \cup (0,1] \ .
\]

\nd We also define the examples labelling $l$ over $\ind$ in the following way: for $a \in \ind$
\begin{equation*}
l(a) = 
\begin{cases*}
1 & if $(a,1) \in \E^+$ \\
-1 & else \ .
\end{cases*}
\end{equation*}

\nd Finally, for a weak hypothesis $h_i$, we determine also the \emph{error} of $h_i$ \wrt~a distribution $\myvec{w}$ as
\[
\epsilon(\myvec{w}) = \sum_{a \in \ind} w_a \cdot \delta(h_i(a),l(a)) \cdot h_i(a) \ ,
\]
\nd where $\delta(x,y) \in \{0,1\}$ is defined as ($x\in \{-1\} \cup (0,1],y \in \{-1,1\}$)
\[
\delta(x,y) = 
\begin{cases*}
1 & if $x \cdot y <0 $\\
0 & else \ .
\end{cases*}
\]
\nd Note that $\delta(h_i(a),l(a))$ determines whether there is a \emph{disagreement} between the sign of  $h_i(a)$ and $l(a)$.

\begin{algorithm}
\begin{algorithmic}[1]
\Require KB $\K$, training set $\E$, target concept name $T$, number of iterations $n$ 
\Ensure Hypothesis $h$ as by  Eq.~\ref{eqnhyp}.
\State $h \gets \emptyset$;
\State $\ind \gets \indkb$;
\State $\myvec{w}_1 \gets \myvec{u}$; \Comment{Initialize the weight distribution over $\ind$} 
\State // Main boosting loop
\For{$i=1$ to $n$} 
\State $h_i \gets$ \Call{FuzzyWeakLearner}{$\KB$, $T$, $\E$, $\myvec{w}_i$}; \Comment{Weak learner $h_i$, \ie~set of axioms $C_{ij} \impc T$}
\If{$\epsilon(\myvec{w}_i) \geq 0.5$} {\bf break}; \Comment{If weak learner error is not below $0.5$ then break the loop} \EndIf
\State $h_i^\star \gets \max_{a \in  \ind} |h_i(a)| $; \Comment{$h_i^\star $ is the maximal value of $h_i$ over $\ind$}
\State $\mu_i \gets \frac{1}{h_i^\star} \sum_{a \in \ind} w_{i,a} \cdot l(a) \cdot h_i(a) $; \Comment{$\mu_i$ is the \emph{normalized} margin of $h_i$ 
\wrt~$\ind$}
\State $ \alpha_i \gets \frac{1}{2h_i^\star} \cdot \ln \frac{1+ \mu_i}{1-\mu_i}$; \Comment{$\alpha_i$ is the weight of classifier $h_i$ in the ensemble} 
\ForAll{$a \in \ind$} \Comment{Update the weight distribution} 
\State $w_{i+1, a} \gets w_{i+1, a} \cdot \left(\frac{1 - (\mu_i \cdot l(a) \cdot h_i(a))/h_i^\star}{1- \mu_i^2}\right) $; 
\EndFor
\State $h \gets h \cup  \{C_{ij} \impc WL_i  \mid C_{ij} \impc T \in h_i, WL_i \mbox{ new}\}$  \Comment{Update  hypothesis according to Eq.~\ref{eqnhyp}.} 
\EndFor

\State // Build now the final classifier ensemble
\State $\phi_T \gets \alpha_1 \cdot WL_1 + \ldots + \alpha_n \cdot WL_n \impc T$; 
\State $h \gets h \cup \{\phi_T \}$;
\State \Return $h$;

\end{algorithmic}
\caption{\fuzzyowladaboost} \label{alg:owladaboost}
\end{algorithm}

Then, the \fuzzyowladaboost~algorithm calling iteratively a weak learner is shown in \refalgo{alg:owladaboost}, which we comment briefly next.

The algorithm is similar as $\mathbb{R}$eal AdaBoost, except for some context dependent parts.
In Step 2 we initialise the set of individuals $\ind$ to be considered as $\indkb$. Essentially, all individuals will be weighted.
The main loop (Steps 5 - 13) is similar to $\mathbb{R}$eal AdaBoost with the particularity that  Step 6 we invoke our  weak learner that is assumed to return a set  
$h_i = \{ C_{i1} \impc T, \ldots, C_{ik_i} \impc T \}$ of fuzzy $\elbool$ GCIs.
In Step 7 we have a case that causes a break of the main loop. In fact, an implicit condition of boosting is that the  error of a weak learner should be below $0.5$.  That is, the weak hypothesis should be better than random guess.

%
%
In Step 13 we update the hypothesis $h$ with the weak hypothesis, while
%
%
 in Steps 15 - 16 we build the final classifier ensemble and add it to the hypothesis.
%
%

\subsection{The Weak Learner  \foildlw} \label{sect:wla}

\nd We next describe the weak learner we employ here. Specifically, we will use  a \foildl~\cite{Lisi13,Lisi13a,Lisi15} like weak learner, which however needs to be adapted to our specific setting 
(see  Algorithm~\ref{alg:wfoilsets}).

\begin{algorithm}
\begin{algorithmic}[1]
\Require KB $\K$, target concept name $T$, training set $\E$,  weight distribution $\myvec{w}$, confidence threshold $\theta \in [0,1]$, non-positive coverage percentage $\eta \in [0,1]$
\Ensure A weak hypothesis, \ie~ a set $h = \{ C_{1} \impc T, \ldots, C_{k} \impc T \}$ of fuzzy $\elbool$ GCIs

	\State $h \gets \emptyset, Pos \gets \calE^+, \phi \gets \topc \impc T$; 
	\State //Loop until no improvement
	\While{($Pos \neq \emptyset$) {\bf and} ($\phi \neq  \mathbf{null}$)} 
		\State $\phi \gets \Call{Learn-One-Axiom}{\KB, T, \E, \myvec{w}_i, Pos, \theta, \eta}$; // Learn one fuzzy $\elbool$ GCI of the form $C \impc T$ 
		\If{$\phi \in h$} // axiom already  learned
			\State $\phi \gets \mathbf{null}$; 
		\EndIf	
		\If{$\phi \neq  \mathbf{null}$}
			\State $h \gets h \cup \{\phi\}$; // Update weak hypothesis
			\State $Pos_\phi \gets \{ \tuple{a,1} \in \calE^{+} \mid \bed{\K \cup \{\phi\}}{T(a) > 0} \}$; // Positives covered by $\phi$
			\State $Pos \gets Pos \setminus Pos_\phi $; // Update positives still to be covered
		\EndIf	
	\EndWhile
\State \Return $h$;
\end{algorithmic}
\caption{\foildlw~(Weak Learner)}
\label{alg:wfoilsets}
\end{algorithm}

\nd In general terms the weak learning algorithm, called \foildlw,  follows a so-called \emph{sequential covering} learning approach. That is, one carries on inducing GCIs until all positive examples are covered or nothing new can be learned. When an axiom is induced (see Step 4 in Algorithm~\ref{alg:wfoilsets}), the positive examples still to be covered are updated (Step 9 and 10).

In order to induce an axiom (Step 4),  \textsc{Learn-One-Axiom} is invoked (see Algorithm~\ref{alg:wfoilOne}), which in general  terms operates as follows:
\begin{enumerate}
\item start from concept $\topc$;
\item apply a refinement operator to find more specific fuzzy $\elbool$ concept description candidates;
\item exploit a scoring function to choose the best candidate;
\item re-apply the refinement operator until a good candidate is found; 
\item iterate the whole procedure until a satisfactory coverage of the positive examples is achieved.
\end{enumerate}

\nd We briefly detail the steps of  \textsc{Learn-One-Axiom}.

\vspace{1ex}
\nd {\bf Computing fuzzy datatypes.}
\nd For a numerical datatype $s$, we coinsider \emph{equal width triangular partitions} of values $V_s = \{ v \mid   \KB \models \cass{a}{\some s.=_{v}} \}$  into a finite number of fuzzy sets ($3,5$ or $7$ sets), which is identical to~\cite{Lisi13,Lisi15,Straccia15}  (see, \eg~Fig.~\ref{partfuzzytrz}). However, we additionally,  allow also the use of the C-means fuzzy clustering algorithm over $V_s$, where the fuzzy membership function is a triangular function build around the centroid of a cluster. Note that C-means has not been considered in~\cite{Lisi13,Lisi15,Straccia15}.\footnote{Specifically, C-means has not been considered so far in fuzzy GCI learning.}

\vspace{1ex}
\nd {\bf The refinement operator.}
The refinement operator  we employ is the same as in~\cite{Lisi13,Lisi13a,Lisi13b,Straccia15} except  that now we add the management of boolean values as well. Essentially, the refinement operator takes as input a concept $C$ and generates new, more specific concept description candidates $D$ (\ie, $\KB \models D \impc C$). For the sake of completeness, we recap the refinement operator here. Let $\KB$ be a knowledge base, ${\bf A}_\KB$ be the set of all atomic concepts in $\KB$, ${\bf R}_\KB$ the set of all object properties in $\KB$, ${\bf S}_\KB$ the set of all numeric datatype properties in $\KB$, ${\bf B}_\KB$ the set of all boolean datatype properties in $\KB$  and $\mathcal{D}$ a set of (fuzzy) datatypes. The refinement operator $\rho$ is shown in Table~\ref{tab:refinement}. 
\begin{table*}
\caption{Downward Refinement Operator.} \label{tab:refinement}
\footnotesize
\[
\rho(C)=\left\{
\begin{array}{lcl}
{\bf A}_\KB \cup \{ \some r.\topc \; | \; r \in {\bf R}_\KB \} \cup \{ \some s.d \; | \; s \in {\bf S}_\KB, d\in \mathcal{D} \}  \cup \\
\hspace{3cm}\{ \some s.=_b,\; | \; s \in {\bf B}_\KB, b \in \{ \mathbf{true},   \mathbf{false}\}\} 
& \mbox{if} & C=\topc \\

\{ A' \; | \; A'\in {\bf A}_\KB, \KB \models A' \impc A \} \cup \{ A\andc A'' \; | \; A''\in\rho(\topc)\}
&  \mbox{if}  & C=A\\

\{ \some r.D'\; | \;D'\in\rho(D) \} \cup \{ 
(\some r.D) \andc D'' \; | \; D''\in \rho(\topc) \}
&  \mbox{if}  & C=\some r.D, r\in{\bf R}_\KB\\

\{ (\some s.d)\andc D \; | \; D\in \rho(\topc) \}
&  \mbox{if}  & C=\some s.d, s\in{\bf S}_\KB,d\in\mathcal{D}\\

\{ (\some s.=_b) \andc D \; | \; D\in \rho(\topc) \}
&  \mbox{if}  & C=\some s.=_b, s\in{\bf B}_\KB,b\in\{\mathbf{true,false}\}\\

\{ C_1 \andc ... \andc C_i ' \andc ... \andc C_n \; | \; i=1,...,n,C_i '\in \rho(C_i)\}
&  \mbox{if}  & C=C_1 \andc ... \andc C_n\\
\end{array}
\right.
\]
\end{table*}

\vspace{1ex}
\nd {\bf The scoring function.}
The scoring function we use to assign a score to each candidate hypothesis is essentially a weighted \emph{gain} function, similar to the one employed in~\cite{Lisi13,Lisi13a,Lisi13b,Straccia15} and implements an information-theoretic criterion for selecting the best candidate at each refinement step.
Specifically, given a fuzzy $\elbool$ GCI $\phi$ of the form $C \impc T$ chosen at the previous step, a KB  $\KB$, a set of individuals $\mathsf{I}$,  a weight distribution  $\myvec{w}$ over $\ind$, a set of positive examples $Pos$ still to be covered and a candidate fuzzy $\elbool$ GCI $\phi^\prime$ of the form $C^\prime \impc T$, then
\begin{equation}\label{eq:gain}
	gain(\phi^{\prime}, \phi,\myvec{w}, \ind, Pos) = p \ast (log_2(cf(\phi^{\prime},\myvec{w},\ind, Pos)) - log_2(cf(\phi,\myvec{w},\ind, Pos))) \ , 
\end{equation}

\nd where $p = |C^\prime \andc C |_\KB^{\myvec{w},\inds{Pos}}$ is the weighted cardinality of positive examples in $Pos$ covered by $\phi$ that are still covered by $\phi^{\prime}$, and 
\begin{equation}\label{wcfpos}
cf(D \impc T, \myvec{w},\ind, Pos) =  \frac{|D|_\KB^{\myvec{w},\ind_{Pos}}}{|D|_\KB^{\myvec{w},\ind}} \ .
\end{equation}


\nd  Please note that in Eq.~\refeq{wcfpos}, about the confidence degree of $D \impc T$, the numerator is calculated 
\wrt~the positive examples still to be covered, \ie~all instances of $T$ that are in $Pos$ and are instances of $D$. In this way,  \textsc{Learn-One-Axiom} is somewhat guided towards positives not yet covered  by the weak learner learned so far by \foildlw.
Note also that the gain is positive if the confidence degree increases.

\vspace{1ex}
\nd {\bf Stop criterion.}
 \textsc{Learn-One-Axiom}~stops when the confidence degree is above a given threshold $\theta \in [0,1]$, or no GCI can be found that does not cover any negative example (in $\calE^-$) above a given percentage. 


\vspace{1ex}
\nd {\bf The \textsc{Learn-One-Axiom}~algorithm.}
The  \textsc{Learn-One-Axiom}~algorithm is defined in \refalgo{alg:wfoilOne}, which we comment briefly as next. Steps 1 - 3 are simple initialisation steps.
Steps 5 - 21 are the main loop from which we may exit in case there is no improvement (Step 16), and the confidence degree of the so far determined GCI 
is above a given threshold or it does not cover any negative example above a given percentage (Step 18). Note that the latter case guarantees soundness of  the weak learner if this percentage is set to $0$. In Step 8 we determine all new refinements, which then are scored in Steps 10 - 15 in order to determine the one with the best gain. At the end of the algorithm, once we exit from the main loop, the best found GCI is returned (Step 22).

\begin{algorithm}
\begin{algorithmic}[1]
\Require KB $\K$, target concept name $T$, training set $\E$,  weight distribution $\myvec{w}$, $Pos$ set of positive examples still to be covered, confidence threshold $\theta \in [0,1]$, non-positive coverage percentage $\eta \in [0,1]$
\Ensure A fuzzy $\elbool$ GCI of the form $C \impc T$
	\State $\ind \gets \indkb$;
	\State $C \gets \topc$;  \Comment{Start from $\top$}
	\State $\phi \gets C \impc T$;
	\State //Loop until no improvement
	\While{$C \neq \mathbf{null}$} 
		\State $C_{best} \gets C$;
		\State $maxgain\gets 0$;
		\State $\calC \gets \rho(C)$;  \Comment{Compute all refinements of $C$}
		\State // Compute the score of the refinements and select the best one
		\ForAll{$C^\prime \in\calC$}
			\State $\phi^\prime \gets C^\prime \impc T$;
			\State $gain\gets gain(\phi^{\prime}, \phi,\myvec{w}, \ind, Pos) $;
			\If{$(gain > maxgain)$ \textbf{and} $(cf(\phi',\myvec{w},\ind, Pos) >  cf(\phi,\myvec{w},\ind, Pos))$}
				\State $maxgain\gets gain$;
				\State $C_{best}\gets C'$;
			\EndIf
		\EndFor
		\If{$C_{best}=C$}  \Comment{No improvement}
			\State //Stop if confidence degree above threshold or no negative coverage below threshold
			\If{$(cf(C_{best} \impc T,\ind) \geq \theta)$ \textbf{and} $(\frac{\ceil{C_{best}}_{\KB}^{\inds{\E^{-}}}}{|\inds{\E^{-}}|} \leq \eta)$}   \textbf{break}; 		
                 \EndIf{}
                 \State // Manage backtrack here, if foreseen
                 \EndIf{}
		\State $C\gets C_{best}$;
		\State $\phi \gets C \impc T$;
	\EndWhile
\State \Return $\phi$;
\end{algorithmic}
\caption{\textsc{Learn-One-Axiom}}
\label{alg:wfoilOne}
\end{algorithm}

\begin{remark}
As for \foildl~(and \pfoildl), the weak learner \foildlw~also allows to use  a backtracking mechanism (Step 19), which, for ease of presentation,  we omit to include. The mechanism is exactly the same as for the \pfoildl-learnOneAxiom described in~\cite[Algorithm 3]{Straccia15}. Essentially, a stack of  top-$k$ refinements is maintained, ranked in decreasing order of the confidence degree from which we pop the next best refinement (if the stack is not empty) in case no improvement has occurred. $C_{best}$ becomes the popped-up refinement.
\end{remark}

%
%
%


\section{Evaluation}
\label{sec:eval}

\nd We have implemented the algorithm within the \emph{FuzzyDL-Learner}\footnote{\url{http://www.umbertostraccia.it/cs/software/FuzzyDL-Learner/}.} system and evaluated it over a set of (crisp) OWL ontologies. All the data and implementation can be downloaded from the FuzzyDL-Learner home page. 

\subsection{Setup}
\nd A number of OWL ontologies from different domains have been selected as illustrated in Tables~\ref{tab:onto} and~\ref{tab:uci}.  A succinct description of them is provided in Appendix~\ref{ontodesc}.
Note that the ontologies in Table~\ref{tab:uci} are not available as OWL 2 ontologies but only as csv format. Therefore, we have translated them from the csv format according to the procedure shown in Appendix~\ref{ucisec}. 
Furthermore, note that the ontologies in Table~\ref{tab:uci} are taken from the well-known \emph{UC Irvine Machine Learning Repository}~\cite{Dua:2019}. While evaluating ontology-based learning algorithms is untypical on  numerical datatype properties\footnote{To the best of our knowledge, we are unaware of any evaluation of ontology-based methods on those data sets.}, we believe it is interesting to do so as an important ingredient of our algorithm is the use of fuzzy concrete datatype properties.


%
\begin{table*}
\caption{Facts about the ontologies of the experiment.}
\label{tab:onto}
{\tiny
\begin{center}{
\begin{tabular}{l||ccccccccc} \hline
\bf{ontology}  & \bf{DL} & \bf{class.}   & \bf{obj. prop.}   & \bf{data. prop.} & \bf{ind.} & \bf{target} $T$ & pos & neg & dth/cj/$\eta$  \\ \hline
FamilyTree & $\mathcal{SROIF(D)}$ & 22   & 52   & 6 & 368  & Uncle  & 46 & 156  & 1/5/0  \\ \hline 
Hotel & $\mathcal{ALCOF(D)}$    & 89   & 3 & 1 & 88 & Good\_Hotel   & 12 & 11 & 1/5/0  \\ \hline 
Moral & $\mathcal{ALC}$ & 46   & 0   & 0 & 202  & ToLearn\_Guilty & 102 & 100 &1/5/0  \\ \hline 
SemanticBible (NTN) & $\mathcal{SHOIN(D)}$ & 51   & 29   & 9 & 723 & ToLearn\_Woman & 46 & 3 & 1/5/0  \\ \hline 
UBA & $\mathcal{SHI(D)}$ &  44   &  26   & 8 & 1268 & Good\_Researcher  & 22 & 113 & 1/5/0  \\ \hline 
WineOnto &  $\mathcal{SHI(D)}$  & 178   & 15   &  7 & 138  & ToLearn\_DryWine  & 15 & - &1/5/0  \\ \hline 
Pair50 &  $\mathcal{ALC}$  & 3   & 6   & 0 & 311  & ToLearn  &  20 & 29 &2/5/0  \\ \hline 
Straight &  $\mathcal{ALC}$  & 3   & 6   & 0 & 347  & ToLearn  &  4 & 50 & 3/5/1.0  \\ \hline \hline
Lymphography &  $\mathcal{ALC}$  & 50   & 0   &  0 & 148  & ToLearn  &  81 & 67 &1/5/1.0  \\ \hline
Mammographic &  $\mathcal{ALC(D)}$  & 20   & 3   &  2 & 975  & ToLearn  &  445 & 516 &3/5/1.0  \\ \hline 
Pyrimidine &  $\mathcal{ALC(D)}$  & 2   & 0   &  27 & 74  & ToLearn  &  20 & 20 &1/5/1.0  \\ \hline 
Suramin &  $\mathcal{ALC(D)}$  & 47   & 3   &  1 & 2979  & ToLearn  &  7 & 10 & 3/5/1.0  \\ \hline \hline
   \end{tabular}}
\end{center}
}
\end{table*}

\begin{table}
\caption{Datasets considered from the UCI ML Repository.} \label{tab:uci}
\begin{center}
{\scriptsize
\begin{tabular}{c||ccccc} \hline
dataset & instances & attributes  & target $T$ & pos & dth/cj/$\eta$ \\ \hline
Iris & 151 & 4 & 
\begin{tabular}{c}
Iris-setosa \\
Iris-versicolor \\
Iris-virginica \\
\end{tabular} &
\begin{tabular}{c}
 51\\
50 \\
 50 \\
\end{tabular} 
& 1/5/1.0 \\ \hline
Wine & 178 & 13 & $1,2,3$ & $59, 71, 48$ & 1/5/1.0 \\ \hline
Wine Quality & 4898 & 12  & GoodRedWine & $18$ & 1/5/1.0 \\ \hline
\end{tabular}
} 
\end{center}
\end{table}

For each ontology $\KB$ a meaningfull target concept has been selected such that the conditions of the learning problem are satisfied. We report also the DL the ontology refers to, the number of concept names, object properties, datatype properties and individuals in the ontology. We also report the maximal nesting depth (dth.), maximal number of conjuncts (cj.) and maximal percentage of false positives ($\eta$) during the  learning phase. 
%
The number $n$ of iterations of \fuzzyowladaboost~is set to $10$.\footnote{We tried also for $n > 10$ and did not notice positive effects. In fact, at some point the weak learner is unable to learn new rules given the weight distribution.} 
We did not consider backtracking. Nevertheless, all configuration parameters for each run are available from the downloadable data.


We will consider the following effectiveness measures (see also~\cite{Straccia15} for similar measures), which we report here for clarity  to avoid ambiguity. 
Specifically, consider the classifier ensemble $h$ returned  by \fuzzyowladaboost~and let us assume to have added it to the KB $\K$.
Then we consider the following fuzzy measures, were their well-known crisp variants~\cite{Baeza99} (in the denotation we omit the $f$ subscript) are obtained by replacing in the equations below  the cardinality function $|\cdot|_\KB^{\indkb}$ (see Eq.~\ref{card}) with the crisp cardinality function $\ceil{\cdot}_\KB^{\indkb}$.

\begin{description}
\item[Fuzzy True Positives:] denoted $TP_f$, is defined as  
\begin{equation} \label{fTP}
TP_f =  |T |_{\KB}^{\inds{\E^+}} \ ,
\end{equation}

\item[Fuzzy False Positives:] denoted $FP_f$, is defined as  
\begin{equation} \label{fFP}
FP_f =  |T|_{\KB}^{\inds{\E^-}} \ ,
\end{equation}

\item[Fuzzy True Non-Positive:] denoted $TNP_f$, is defined as  
\begin{equation} \label{fTNP}
TNP_f =  |\inds{\E^-}| - FP\ ,
\end{equation}

\item[Fuzzy False Non-Positive:] denoted $FNP_f$, is defined as  
\begin{equation} \label{fFNP}
FNP_f =  |\inds{\E^+}| - TP\ ,
\end{equation}

\item[Fuzzy Precision:] denoted $P_f$, is defined as  
\begin{equation} \label{prec}
P_f =   \frac{TP_f}{|T|_\KB^{\inds{\E}}} \ ,
\end{equation}

\item[Fuzzy Recall:] denoted $R_f$, is defined as  
\begin{equation} \label{rec}
R_f =  \frac{TP_f}{|\inds{\E^+}|} \ ,
\end{equation}

\item[Fuzzy $F1$-score:] denoted $F1_f$, is defined as
\begin{equation*}
F1_f = 2 \cdot \frac{P_f\cdot R_f}{P_f + R_f} \ , 
\end{equation*}


\item[Mean Squared Error:] denoted $MSE$, is defined as  
\begin{equation*}
MSE = \frac{1}{|\inds{\E}|} \cdot \sum_{a \in \inds{\E}} (h(a) - l(a))^2 \ .
\end{equation*}
%
\end{description}


\nd Concerning other parameter settings, we
\begin{itemize}
\item varied the number of fuzzy sets ($3, 5$ or $7$). For C-means, we fixed the hyper-parameter to the default $m=2$, the threshold to $\epsilon = 0.05$ and the number of maximum iterations to $100$; and
\item varied the confidence threshold $\theta \in \{0.34, 0.64, 0.94, 1.0\}$.
\end{itemize}
\nd Therefore,  we considered a total of $12$ different parameter configurations. 

For each parameter configuration, a stratified $k$-fold cross validation design\footnote{Stratification means here that each fold contains roughly the same proportions of positive and non-positive instances of the target class.} was adopted (specifically, $k=5$) to determine the average of the above described performance indices.
For each measure, the (macro) average value over the various folds has been considered.
In all tests, we have that $\inds{\E} = \indkb$ and that there is at least one positive example in each fold, while the other examples of a fold have been randomly been selected.
 For each fold, all assertions involving testing examples have been removed from a given ontology, thus restricting the training phase to training examples only. 
We considered also the extreme case in which the whole set $\E$ is used for both training and testing. This case has been considered for those ontologies with few positive examples for which $k$-fold cross validation is not meaningful and also for the task aiming at ``explaining" the target \wrt~the given data set. This case is indicated with  $\star$ in Table~\ref{tab:allres}.

As baseline, we considered an improved version of  \foildl~\wrt~the one published in~\cite{Lisi13,Lisi13a,Lisi15}. Roughly, \foildl, as \foildlw, learns iteratively rules. At each iteration $i$,  \foildl~learns one fuzzy $\elbool$ GCI of the form
$\fuzzyg{C_i  \impc   T}{\m_i}$
by invoking a similar procedure as \textsc{Learn-One-Axiom}, where however
\begin{itemize}
\item $d_i$ is the confidence degree of $C_i  \impc   T$; and 
\item the weight distribution $\myvec{w}$ is roughly as follows: if a positive instance $a$ has already been covered by the rules learned so far, then $w_a=0$ (this is the same as to say to remove the covered positive instances from the next iteration). The weight of the other instances is determined according to a \emph{uniform} distribution.
\end{itemize}
\nd  At the end, the final hypothesis of \foildl~is of the form (\cf~Eq.~\ref{eqnhyp})
\begin{equation} \label{hypofoilA}
\begin{array}{l}
\fuzzyg{C_1  \impc   T}{d_1} \\
\hspace{1cm} \vdots  \\
\fuzzyg{C_n  \impc   T}{d_n} \ .  \\
\end{array}
\end{equation}

\nd Therefore, there is a notable difference among  \fuzzyowladaboost~and \foildl~ in \ii{i} the way the instance distribution is set up at each iteration; \ii{ii} how the weight of each rule is determined (the $\alpha_i$ in \fuzzyowladaboost~versus the confidence $d_i$ in \foildl); and \ii{iii} how the final hypothesis is build (linear combination in \fuzzyowladaboost~versus `$\max$' aggregation in \foildl).

In the result Table~\ref{tab:allres}, for a given KB $\K$, a given algorithm (\fuzzyowladaboost~or \foildl) and a given clustering method (uniform $u$ or C-means $c$),
we report only the effectiveness measures for the configuration $(\theta, \mbox{fs})$\footnote{Recall that $\theta \in \{0.34, 0.64, 0.94, 1.0\}, \mbox{fs} \in \{3,5,7\}$.} with the highest score of 
%
\begin{equation} \label{eqbest}
fF1F1 = F1_f \cdot F1 \ ,
\end{equation}

\nd \ie~a compromise (Pareto optimal solution)  among fuzzy $F1$  and crisp $F1$, as, more often than not, the best fuzzy $F1$  and best crisp $F1$ values do not relate to the same configuration.\footnote{In case of a tie, we adopt the following priorities: lowest $\theta$ and then lowest number of partitions.}

\begin{example} \label{exh}
\nd We provide here an example of learned rule set (in Machester OWL syntax)  via \fuzzyowladaboost~applied to the Wine dataset (see Table~\ref{tab:uci}) and target class $2$ (considering the best run).

%
%

{\tiny
\begin{verbatim}
# Weak Learner WL1
(Alcohol some Alcohol_VL) and (Hue some Hue_H) and (MalidAcid some MalidAcid_VVL) SubClassOf WL1
(Ash some Ash_L) and (ColorIntensity some ColorIntensity_VVL) SubClassOf WL1
(ColorIntensity some ColorIntensity_VVL) and (Proline some Proline_L) SubClassOf WL1
(ColorIntensity some ColorIntensity_VVL) and (Proline some Proline_VVL) SubClassOf WL1
(Magnesium some Magnesium_VVL) and (NonFlavonoidsPhenols some NonFlavonoidsPhenols_VVH) and (Proanthocyanins some Proanthocyanins_F) SubClassOf WL1
(MalidAcid some MalidAcid_VVL) and (Proline some Proline_VVL) SubClassOf WL1

# Weak Learner WL2
(Alcohol some Alcohol_VVL) and (ColorIntensity some ColorIntensity_VVL) SubClassOf WL2
(Ash some Ash_VVL) and (Flavonoids some Flavonoids_L) SubClassOf WL2
(ColorIntensity some ColorIntensity_VVL) and (Proline some Proline_VVL)) SubClassOf WL2

# Weak Learner WL3
(ColorIntensity some ColorIntensity_VVL) SubClassOf WL3

# Real Adaboost aggregation
1.199 * WL1 + 0.544 * WL2 + 0.272 * WL3 SubClassOf 2

# Fuzzy datatypes (C-Means)
Alcohol_VVL				left-shoulder	11.81	12.29	
Alcohol_VL				triangular		11.79	12.27	12.58
Hue_H					triangular		1.00		1.08		1.21
MalidAcid_VVL				left-shoulder	1.27		1.67	
Ash_VVL	 				left-shoulder	1.88		2.11	
Ash_L					triangular		2.12		2.25		2.35
ColorIntensity_VVL			left-shoulder	2.66		3.64	
Proline_L					triangular		532.73	657.29	811.52
Proline_VVL				left-shoulder	424.58	532.73
Magnesium_VVL			left-shoulder	85.39	89.97
NonFlavonoidsPhenols_VVH	right-shoulder	0.50		0.56	
Proanthocyanins_F	 		triangular		1.40		1.64		1.94
Flavonoids_L				triangular		1.35		1.76		2.21
\end{verbatim}
}

\qed
\end{example}

\begin{table}
\caption{Results table.} \label{tab:allres}
\begin{center}
\includegraphics[scale=0.425]{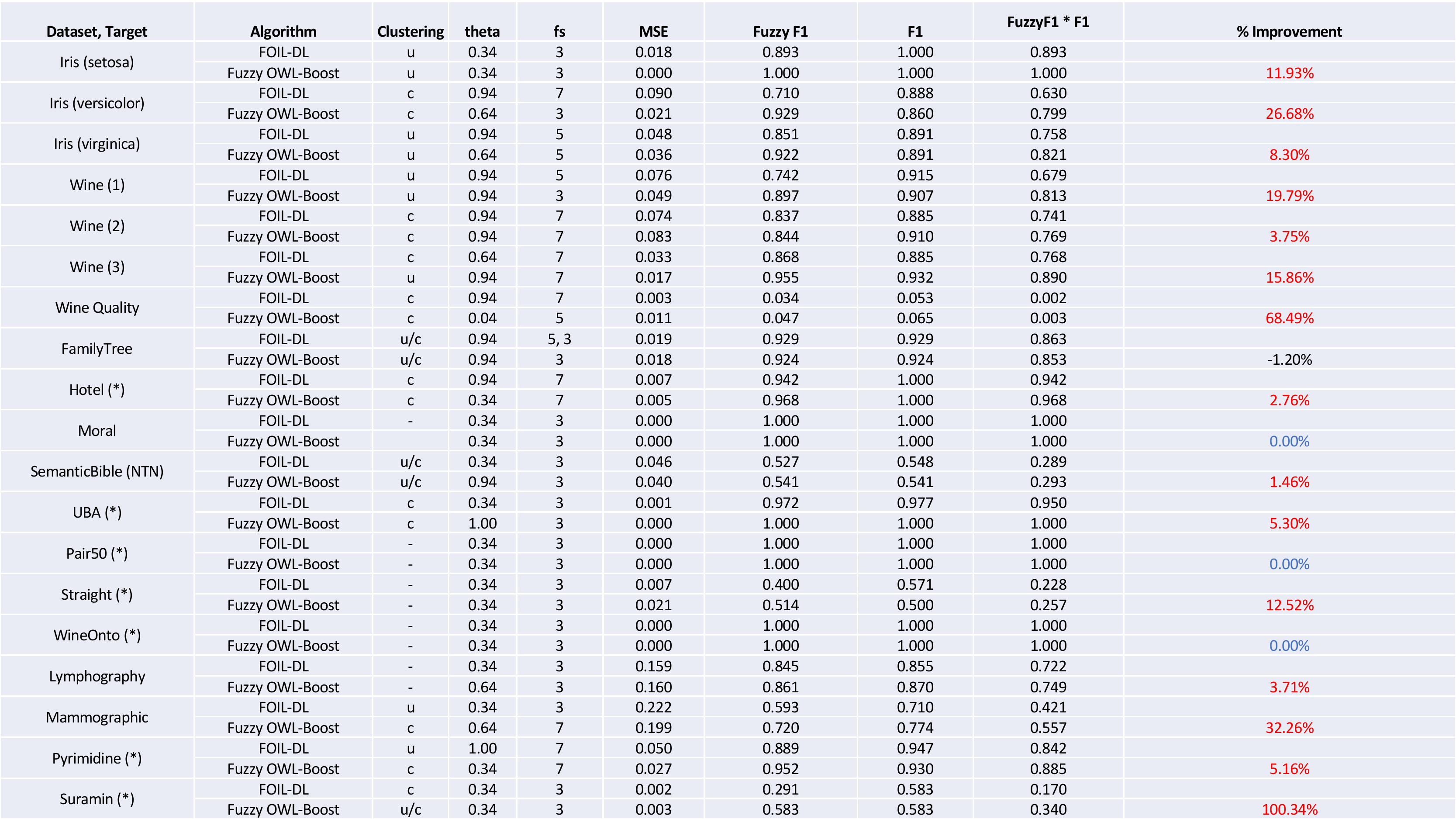}
\end{center}
\end{table}

\subsection{Discussion}

\nd We now discus the results in Table~\ref{tab:allres}. We report in red the percent improvement of \fuzzyowladaboost, relative to the measure $fF1F1$ (see Eq.~\ref{eqbest}), over our baseline \foildl.

\paragraph{Uniform  vs.~C-Means fuzzy datatype construction} To start with, without going to much into it as it is not the main of this work, not surprisingly C-means behaves better than the `uniform' ($u$) approach (14 wins vs. 9).\footnote{We do not count ties.}  Moreover, concerning the number of fuzzy set partitions, there is no clear indication about which choice between $3$ or $7$ partitions is the better way to go. The choice seems dependent on the dataset. Apparently, if $3$ partitions are not enough, then one may go for $7$ as likely the dataset may require a more fine grained approach.
Of course,  the results of C-means may further be improved by optimising its parameters. Nevertheless, the uniform approach performed surprisingly well, despite its simplicity.

A more in depth investigation will be subject of future work in which we will consider some more options for fuzzy set construction and focus on its impact on the overall effectiveness.

%
%

\begin{figure}
\begin{center}
\includegraphics[scale=0.5]{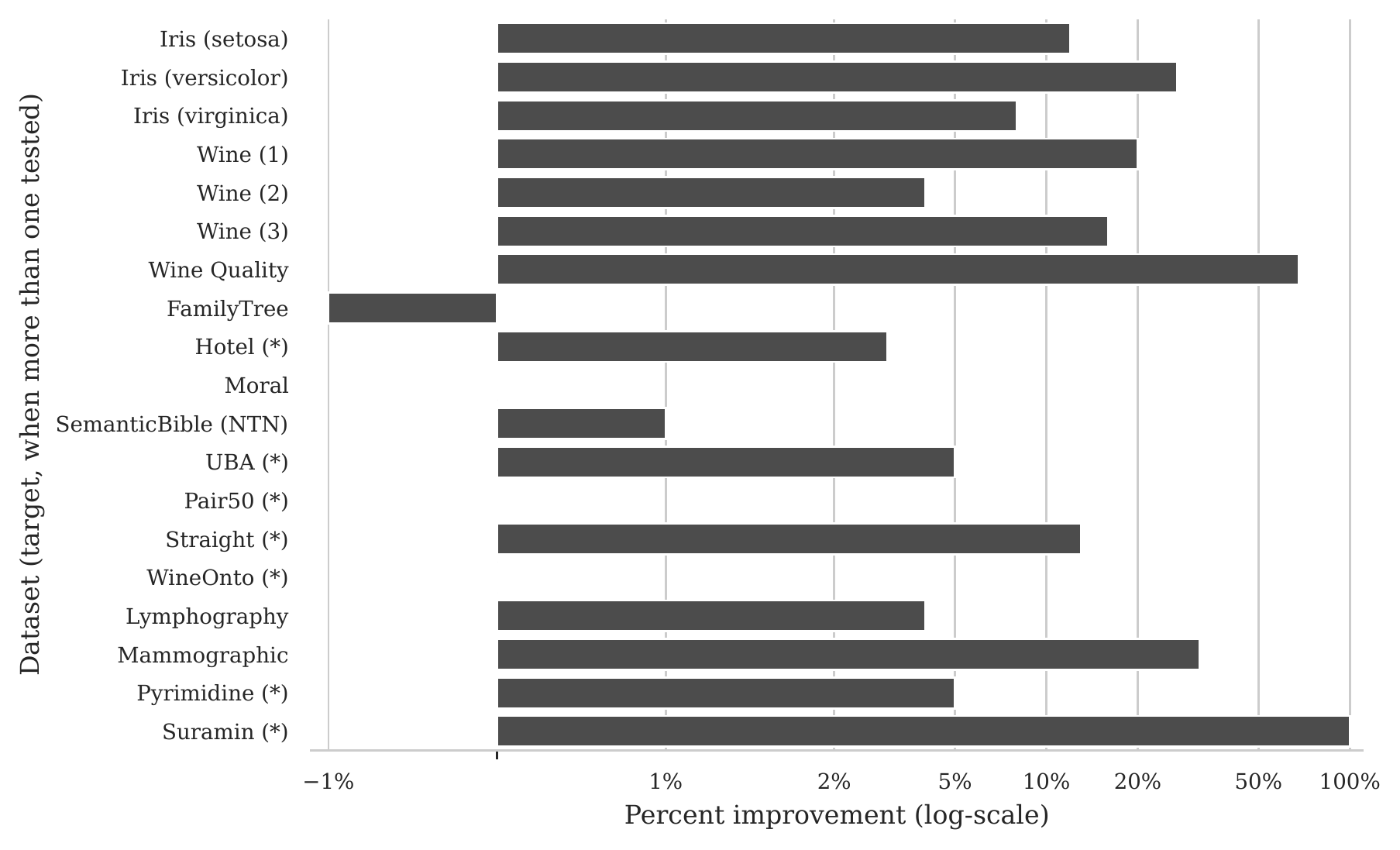}
\end{center}
  \caption{Improvement of \fuzzyowladaboost~over \foildl~according to the $fF1F1$ (Eq.~\refeq{eqbest}) measure.}
  \label{figbest}
\end{figure}

\paragraph{\fuzzyowladaboost~vs.~\foildl} 
%
%
%
\nd It appears evident from the results in Table~\ref{tab:allres} (see also~\reffig{figbest}) that \fuzzyowladaboost~performs generally better than \foildl~(15 wins vs.~1), with 3 ties and 1 loss.
Concerning the ties, note that the value of $fF1F1$ is $1.0$ for both \fuzzyowladaboost~and \foildl~and, thus, there was no margin for improvement for \fuzzyowladaboost. The only loss was for the FamilyTree dataset, though, the difference is small ($-1.20\%$).

The average improvement of \fuzzyowladaboost~over all runs is $16.69\%$.
This is essentially due to the average improvement \wrt~the fuzzy F1 measure, which is $15.16\%$ (15 wins vs.~1, 3 ties), while the average improvement for the (crisp) F1 measure is marginal $1.21\%$ (6 wins vs.~7, 6 ties). 

Overall, note also that the average $MSE$ is low ($0.041$) with a slightly advantage for \fuzzyowladaboost~over \foildl~($0.036$ vs.~$0.045$). Nevertheless, there are two outlier datasets (Lymphography and Mammographic) for which there is quite some room for improvement of the $MSE$.

Last but not least, let us mention that  both  \fuzzyowladaboost~and \foildl~do definitely not behave well on the 
WineQuality dataset, which will be the subject of further investigation.\footnote{However, a run on this dataset requires ca.~one week of computation on our hardware 
(Linux OS, with 16GB RAM and Intel Core i9-9900K CPU @ 3.60GHz).}
 
%
%




\section{Conclusions \& Future Work}
\label{sec:conclusions}

\nd In this work, we  addressed the problem of automatically learning fuzzy concept inclusion axioms from OWL 2 ontologies. That is, given a  target class $T$ of an OWL ontology, we address the problem of inducing a fuzzy $\ELW(\D)$ concept inclusion axioms that describe sufficient conditions for being an individual instance  of $T$. In particular,  we have adapted the
$\mathbb{R}$eal AdaBoost~\cite{Nock07} boosting algorithm  to the fuzzy OWL case, by presenting the \fuzzyowladaboost~algorithm.
The main features of our algorithm are essentially the fact that
\ii{i} it generates a set of  fuzzy $\elbool$ inclusion axioms, which are the weak hypothesis, possibly including fuzzy concepts and fuzzy concrete domains; 
\ii{ii} combines them via a weighted sum; and
\ii{iii} all generated fuzzy concept inclusion axioms can  be encoded as \emph{Fuzzy OWL 2} axioms.

We have also conducted an extensive evaluation, comparing \fuzzyowladaboost~with \foildl. Our evaluation shows that \fuzzyowladaboost~is generally better than \foildl~in terms of $fF1F1$ effectiveness (+$16.69\%$ average) over the tested datasets, and that the improvement mainly concerns the fuzzy F1 measure, while effectiveness remains essentially similar for (crisp) F1. Also, the C-means clustering method prevails over the uniform clustering method to build the fuzzy data types. Let us also note that both \fuzzyowladaboost~ (as well as \foildl) generates easy human interpretable hypotheses (see \eg~Example~\ref{exh}).

Last but not least, let us mention that in a previous version of this work, we also considered the use of the \textsc{Learn-One-Axiom} algorithm only (see Algorithm~\ref{alg:wfoilOne}) as weak learner in place of \foildlw~and the softmax function to normalise the weights $\alpha_i$ in the weighted sum construct. However, the results were not really encouraging (\ie~slightly worse) \wrt~\foildl.

Concerning future work, besides investigating about other learning methods, we envisage various aspects worth to be investigated in more detail: 
$(i)$ we would like to make an in depth investigation about the impact of clustering methods for building fuzzy datatypes on the overall effectiveness by considering various alternatives as well, as proposed recently in a Fuzzy Sets and Systems special issue on fuzzy clustering~\cite{fssclustering20}. Moreover, we would like to cover more OWL datatypes than those considered here so far (numerical and boolean) such as strings, dates, etc. possibly in combination with some sub-atomic classical machine learning methods (see, \eg~\cite{ShalevShwartz14});
\ii{ii} another aspect may concern the investigation of the impact of choosing various fuzzy semantics during the learning phase; 
$(iii)$ last but not least, we would like to investigate the computational aspect: so far, for some ontologies, a learning run may take even a week (on the resource at our disposal\footnote{See footnote 18.}). We would like to investigate both parallelisation methods as well as to investigate about the impact, in terms of effectiveness, of efficient, logically sound, but not necessarily complete, reasoning algorithms, such as \emph{structural} DL algorithms.



\section*{Acknowledgment}
\nd This research was partially supported by TAILOR, a project funded by EU Horizon 2020 research and innovation programme under GA No 952215. This work has also been partially supported by the H2020 DeepHealth Project (GA No. 825111).


\appendix
\renewcommand{\thesection}{\Alph{section}}
\setcounter{section}{0}
\section{Brief Description of the Datasets}\label{ontodesc}

\subsection{OWL Ontologies Description}

\nd Find below a brief description about the OWL ontologies in Table~\ref{tab:onto} used in our experiments. Some ontology descriptions can also found in~\cite{Westphal19}.\footnote{See also, \url{https://github.com/SmartDataAnalytics/SML-Bench}}

\begin{description}
\item[FamilyTree.] This is a simple family relationships ontology and associated instances. The description is of the family of Robert Stevens and the intention is to use the minimal of asserted relationships and the maximum of inference. To do this,  role chains, nominals and properties hierarchies have been used. The target is to identify sufficient conditions for being an uncle.

\item[Hotel.] This ontology describes the meaningful entities of a city. Instances are hotels located in the town Pisa and ratings have been gathered from Trip Advisor.\footnote{\url{http://www.tripadvisor.com}} The target is to identify sufficient conditions for being a good hotel, which has been identified as a hotel having a rating above 4.

\item[Moral.]  This ontology is about meaningful  entities involved in the description of guiltiness within a moral  theory of blame scenario. The target is to learn sufficient conditions to be guilty.

\item[SemanticBible (NTN).] \emph{New Testament Names} (NTN) is an ontology describing each named thing in the New Testament, about 600 names in all. Each named thing (an entity) is categorized according to its class, including God, Jesus, individual men and women, groups of people, and locations. These entities are related to each other by properties that interconnect the entities into a web of information.\footnote{\url{http://semanticbible.com/ntn/ntn-overview.html}} The target is to learn sufficient conditions to be a woman.

\item[UBA.] This is a well-known  university ontology for benchmark tests describing meaningfull entities within a university (\eg~universities, departments and the activities that occur at them).\footnote{\url{http://swat.cse.lehigh.edu/projects/lubm/}} The target here is to determine sufficient conditions to be a good researcher.

\item[WineOnto.] This is an ontology about Italian, French and German red and white wines involving the description of, among others, their chemical properties. The target here is to determine sufficient conditions to be a dry wine.

\item[Pair50.] This ontology is about a poker game and the target is to determine whether a player has a pair at hand.

\item[Straight.] This ontology is about a poker game, as the one for Pair50, but the target is now to determine whether one has a straight at hand.

\item[Lymphography.] This ontology is about lymphography patient data and the target is the prediction of a diagnosis class based on the lymphography patient data~\cite{Westphal19}.

\item[Mammographic.] This ontology is about mammography screening data and the target is the prediction of breast cancer severity based on the screening data~\cite{Westphal19}.

\item[Pyrimidine.] This ontology is about pyrimidine data, the target is the prediction of the inhibition activity of pyrimidines and the DHFR enzyme~\cite{Westphal19}.

\item[Suramin.] This ontology is about the description of chemical compounds and the target is to find a predictive description of suramin analogues for cancer treatment.

\end{description}

\subsection{UCI ML Data Sets} \label{ucidesc}
\nd The data sets in Table~\ref{tab:uci} have been taken from the well-known \emph{UC Irvine Machine Learning Repository}~\cite{Dua:2019}. A brief description of the selected data is given below.

\begin{description}
\item[Iris.] The data set contains 3 classes of 50 instances each, where each class refers to a type of iris plant.
The attributes are: 
sepal length in cm,
sepal width in cm,
petal length in cm and
petal width in cm.
The target classes are: Iris Setosa, Iris Versicolour and Iris Virginica.

\item[Wine.] These data are the results of a chemical analysis of wines grown in the same region in Italy but derived from three different cultivars. The analysis determined the quantities of 13 constituents found in each of the three types of wines. The attributes are
alcohol, malic acid, ash, alcalinity of ash, magnesium, total phenols, flavonoids, nonflavonoid phenols, proanthocyanins, color intensity, hue, OD280/OD315 of diluted wines  and proline. The target classes are the three wines $1,2$ and $3$.

\item[Wine Quality.] The  data set is related to red and white variants of the Portuguese ``Vinho Verde" wine. 
The attributes are: 
fixed acidity, volatile acidity, citric acid, residual sugar, chlorides, free sulfur dioxide, total sulfur dioxide, density, pH, sulphates, alcohol and quality (score between 0 and 10).
The target class is GoodRedWine, which is defined as red wines having quality score above 7. 


\end{description}

\section{UCI ML Conversion Algorithm} \label{ucisec}

\nd We considered the well-known \emph{UC Irvine Machine Learning Repository}~\cite{Dua:2019} from which selected some popular datasets with numerical attributes as shown in Table~\ref{tab:uci}.  As anticipated, as the the datasets  in  Table~\ref{tab:uci} are not available as OWL 2 ontologies, we have translated them from a csv format into an OWL 2  ontology in a  simple way that we describe next. 
The method is quite general and can be applied to any other dataset with similar specifications and  a dedicated procedure is available within our implemented learner for future evaluations.

Consider a dataset $D$ with (functional) attributes 
$s_1, \ldots, s_n$ of type $t_1, \ldots, t_n$. Each data record $r$ is of the form
$\tuple{v_1, \ldots, v_n, T}$, where $v_i$ is the value of attribute $s_i$ of type $t_i$, while $T$ is the target class name for record $r$. For instance, for the iris dataset we have attributes
\[
\mathtt{sepal\_length,sepal\_width,petal\_length,petal\_width}
\]
\nd of type
\[
\mathtt{double, double, double, double}
\]
\nd and the first record $r$ is
\[
\tuple{\mathtt{5.1,3.5,1.4,0.2,Iris-setosa}} \ .
\]
\nd The knowledge base $\KB_D$  built to describe the data is as follows. Let $T_D$ be the set of all target class names $T$ occurring in $D$. The set of GCIs in $\K_D$ is 
\begin{equation} \label{tboxuci}
\begin{array}{lcll}
T & \impc & \mathtt{class} &(T \in T_D)\\
\mathtt{class} & \impc & \some s_i.t_i  & (i=1...n) \ .
\end{array}
\end{equation}
\nd Additionally, each data property $s_i$ has been declared as functional.

The set of assertions in $\K_D$ is built in the following way. For each record $r$ of the form $\tuple{v_1, \ldots, v_n, T}$, we create a new individual $a_r$ and add the axioms 
\begin{equation} \label{aboxuci}
\begin{array}{ll}
\cass{a_r}{T} \\
\cass{a_r}{\some s_i.=_{v_i}}  & (i=1...n)
\end{array}
\end{equation}
\nd to $\K_D$. 
 \nd For instance, for the  iris dataset described above, that has three target classes $\mathtt{Iris-setosa}$, $\mathtt{Iris-versicolor}$ and $\mathtt{Iris-virginica}$, the KB contains the axioms
\[
\begin{array}{l}
\mathtt{Iris-setosa} \impc \mathtt{class} \\
\mathtt{Iris-versicolor} \impc \mathtt{class} \\
\mathtt{Iris-virginica} \impc \mathtt{class} \\
\mathtt{class} \impc \some \mathtt{sepal\_length}.\mathbf{double} \\
\mathtt{class} \impc \some \mathtt{sepal\_width}.\mathbf{double} \\
\mathtt{class} \impc \some \mathtt{sepal\_length}.\mathbf{double} \\
\mathtt{class} \impc \some \mathtt{sepal\_width}.\mathbf{double} \\
\cass{a_1}{\mathtt{Iris-setosa}} \\
\cass{a_1}{\some \mathtt{sepal\_length}.=_{5.1}} \\
\cass{a_1}{\some \mathtt{sepal\_width}.=_{3.5}} \\
\cass{a_1}{\some \mathtt{petal\_length}.=_{1.4}} \\
\cass{a_1}{\some \mathtt{sepal\_width}.=_{0.2}} \ .
\end{array}
\]

\nd It is easily verified that the KB $\KB_D$ constructed for each dataset $D$ 
 (i) belongs to the DL $\el(\D)$ extended with functional properties; 
(ii) the number of classes is $|T_D| + 1$; and there are $n$ functional datatype properties.

\section*{References}

{\footnotesize

}

\end{document}